\documentclass[journal,letterpaper]{IEEEtran}
\usepackage[pdftex]{graphicx}
\usepackage[numbers,sort&compress]{natbib}
\usepackage{booktabs}
\usepackage{moreverb}
\usepackage{url}
\usepackage{amsmath}
\usepackage{multicol,lipsum}
\usepackage{mathtools}
\usepackage{cuted}
\usepackage{amsfonts}
\usepackage{multirow}
\usepackage{bm}
\usepackage{subcaption}
\usepackage{color, colortbl}
\usepackage{makecell}

\usepackage{soul}

\usepackage[usenames,dvipsnames]{xcolor}%\usepackage{xcolor,colortbl}
\definecolor{blue_iit}{RGB}{51,51,255}
\usepackage{algpseudocode}
\usepackage{algorithm}
\usepackage{glossaries}
\usepackage[tight]{units}
\usepackage[normalem]{ulem}
\usepackage{cancel}
\definecolor{Gray}{gray}{0.9}
\definecolor{dukeblue}{rgb}{0.0, 0.0, 0.90}
\definecolor{uablue}{rgb}{0.0, 0.2, 0.67}
\usepackage{todonotes}
\usepackage[perpage,multiple]{footmisc}

\renewcommand{\theenumi}{\alph{enumi}}

\usepackage[colorlinks,bookmarksopen,bookmarksnumbered,citecolor=red,urlcolor=red, bookmarks=false]{hyperref} %does not work with IEEEtran!!!
\hypersetup
{
	pdftitle = {Whole-body control for quadrupedal locomotion on challenging terrain},
	pdfauthor = {Shamel Fahmi},
	pdfsubject = {RA-L manuscript},
	pdfkeywords = {whole-body control, legged robots, challenging locomotion, and
		terrain mapping},
	pdftoolbar = true,
	colorlinks = true,
	linkcolor = black,
	citecolor = black,
	urlcolor = black,
}
\newacronym{hyq}{HyQ}{Hydraulically actuated Quadruped}

\newacronym{lf}{LF}{Left-Front}
\newacronym{rf}{RF}{Right-Front}
\newacronym{lh}{LH}{Left-Hind}
\newacronym{rh}{RH}{Right-Hind}

\newacronym{haa}{HAA}{Hip Adduction-Abduction}
\newacronym{hfe}{HFE}{Hip Flexion-Extension}
\newacronym{kfe}{KFE}{Knee Flexion-Extension}

\newacronym{efr}{EFR}{Extended Feasible Region}
\newacronym{imu}{IMU}{Inertial Measurement Unit}
\newacronym{dofs}{DoFs}{Degrees of Freedom}
\newacronym{rt}{RT}{Real Time}

\newacronym{com}{CoM}{Center of Mass}
\newacronym{cop}{CoP}{Center of Pressure}
\newacronym{zmp}{ZMP}{Zero Moment Point}
\newacronym{icp}{ICP}{Instantaneous Capture Point}
\newacronym{cmp}{CMP}{Centroidal Moment Pivot}
\newacronym{grfs}{GRFs}{Ground Reaction Forces}

\newacronym{ls}{LS}{Least Square}
\newacronym{to}{TO}{Trajectory Optimization}
\newacronym{lip}{LIP}{Linear Inverted Pendulum}
\newacronym{slip}{SLIP}{Spring Loaded Inverted Pendulum}
\newacronym{eom}{EoM}{Equation of Motions}
\newacronym{qp}{QP}{Quadratic Program}
\newacronym{sqp}{SQP}{Sequential Quadratic Programming}
\newacronym{mic}{MIC}{Mixed-Integer Convex}
\newacronym{cmaes}{CMA-ES}{Covariance Matrix Adaptation Evolution Strategy}
\newacronym{ara}{ARA*}{Anytime Repairing A*}
\newacronym{pca}{PCA}{Principal Component Analysis}
\newacronym{cpg}{CPG}{Central Pattern Generator}
\newacronym{wbc}{WBC}{Whole-Body Control}

\newacronym{mpc}{MPC}{Model Predictive Control}
\newacronym{ip}{IP}{Iterative Projection}

\newacronym{cwc}{CWC}{Contact Wrench Cone}
\newacronym{awp}{AWP}{Actuation Wrench Polytope}
\newacronym{fwp}{FWP}{Feasible Wrench Polytope}
\newacronym{gws}{GWS}{Grasp Wrench Space}
\newacronym{wfw}{WFW}{Wrench-Feasible Workspace}
\newacronym{fsw}{FSW}{Feasible Solution of Wrench}
\newacronym{FWP}{FWP}{Feasible Wrench Polytope}

\newacronym{awbc}{c$^3$WBC}{Compliant Contact Consistent Whole-Body Control}
\newacronym{swbc}{sWBC}{Standard Whole-Body Control}
\newacronym{c3wbc}{c$^3$WBC}{Compliant Contact Consistent Whole-Body Control}
\newacronym{ste}{TCE}{Terrain Compliance Estimator}
\newacronym{c3}{\texttt{c}$^3$}{compliant contact consistent}

\newacronym{stance}{STANCE}{\textbf{S}oft \textbf{T}errain \textbf{A}daptation a\textbf{n}d \textbf{C}ompliance \textbf{E}stimation}

\newacronym{wbopt}{WBOpt}{Whole Body Optimization}

\newacronym{hc}{HC}{Hunt and Crossley's}
\newacronym{kv}{KV}{Kelvin-Voigt's}

\newacronym{wllsr}{WLLSR}{Weighted Linear Least Squared Regression}

\newacronym{mae}{MAE}{Mean Absolute Tracking Error}

\newacronym{ode}{ODE}{Open Dynamics Engine}

\newcommand{\Rnum}{\mathbb{R}} % Symbol fo the real numbers set

\newcommand{\degree}{\ensuremath{^\circ}}				% define the degree symbol
\DeclareMathOperator*{\argmax}{\arg\!\max}				% argmax
\newcommand{\mat}[1]{\ensuremath{\begin{bmatrix}#1\end{bmatrix}}}	% matrix
\newcommand{\diag}{\ensuremath \mathrm{diag}}
\hypersetup{draft}
\makeglossaries

\usepackage{xcolor}

\newcommand{\eg}{\emph{e.g.,~}}

\newcommand{\ie}{\emph{i.e.,~}}

\definecolor{customred}{rgb}{0.8, 0, 0}
\definecolor{customorange}{rgb}{0.8, 0.33, 0}
\newcommand{\revisedtext}[1]{{\color{black}#1}}
\newcommand{\revisedtextthird}[1]{{\color{black}#1}}

\title{An Efficient Paradigm for Feasibility Guarantees in Legged Locomotion}
\author{Abdelrahman Abdalla$^1$,  Michele Focchi$^1$, Romeo Orsolino$^2$ and Claudio Semini$^1$
	\thanks{$^1$ The authors are with the Dynamic Legged Systems lab, Istituto Italiano di Tecnologia (IIT), Genova, 
		Italy.
		{\tt\small \href{mailto:name.surname@iit.it}{name.surname@iit.it}}}
	\thanks{$^2$ This work was carried out while Orsolino was at the Oxford Robotics Institute, University of Oxford, Oxford, UK.
		{\tt\small \href{mailto:orso.romeo@gmail.com}{orso.romeo@gmail.com}}}}
\usepackage{nccmath}

\begin{document}
	\maketitle
	\thispagestyle{empty}
	\pagestyle{empty}
	
	\begin{abstract}%150-250 word abstract
		Developing feasible body trajectories for legged systems on arbitrary terrains is a challenging task.
		In this paper, we present a paradigm that allows to design feasible \revisedtext{\gls{com} and body} trajectories in an efficient manner.
		\revisedtext{In our previous work \cite{orsolino19tro}, we introduced the notion of the 2D \textit{feasible region}, where static balance and the satisfaction of joint torque limits were guaranteed, whenever the projection of the CoM lied inside the proposed admissible region.
			In this work we propose a general formulation of the \textit{improved feasible region} that guarantees \textit{dynamic} balance alongside the satisfaction of both joint-torque and kinematic limits in an efficient manner.}
		To incorporate the feasibility of the kinematic limits,
		we introduce an algorithm that computes the \textit{reachable region} of the \gls{com}.
		Furthermore, we propose an efficient planning strategy that utilizes the 
		improved feasible region to design feasible \gls{com} and body orientation trajectories.
		Finally, we validate the capabilities of the improved feasible region and the effectiveness 
		of the proposed planning strategy, using simulations and experiments on the 90 kg \gls{hyq} \revisedtext{and the 21 kg Aliengo} robot\revisedtext{s.}
	\end{abstract}
	
	
	\section{Introduction}
	\label{sec:introduction}
	%why planning is useful 
	The central ambition in legged robots development is the ability to traverse unstructured environments.
	This will allow the use of legged robots in difficult applications such as nuclear 
	plants decommissioning, search and rescue missions, and space crater explorations.
	Due to the complexity of the terrain, the demanding payloads, and the variety of obstacles encountered during such operations,
	challenging demands are posed on the robot joints in terms of required actuation efforts and range of motion.
	%feasible trajectories
	Therefore, planning  trajectories that are \textit{feasible}
	becomes crucial for the success of the locomotion task.
	A feasible trajectory in this manuscript is defined to be one that
	fulfills physical constraints in terms of \textit{contact stability}, \textit{joint-torque} and \textit{kinematic} limits.
	
	%numerical optimization
	A powerful tool that is often utilized to devise feasible 
	trajectories is numerical optimization.
	%realtime optimization
	\revisedtext{In} recent years, the availability of increased computational power and the formulation of more efficient algorithms,
	allowed implementations of \revisedtext{optimization-based 
		approaches} that are compatible with
	real-time requirements \cite{Neunert2018, Carpentier2018a}.
	Nonetheless, despite their remarkable achievements, all the proposed approaches employ simplified models that usually avoid considering joint-torque and kinematic limits or perform conservative approximations. 
	
	%heuristics
	On the other hand, heuristic approaches with some 
	or no predictive capabilities were used to successfully address rough terrains through blind locomotion \cite{barasuol13icra} or
	by employing visual feedback to construct (online) the map of the environment \cite{Focchi2020}. 
	Their advantage is their \revisedtext{small} computational
	\revisedtext{effort} that enabled to easily implement them online on a real  \revisedtext{robot}. 
	However, these heuristic approaches fail to provide any
	guarantee on the feasibility of the computed trajectories.
	
	%reduced model 
	Other optimization approaches, employ approximate (\ie reduced) models to reduce the number of states and achieve 
	on-line re-planning in a \gls{mpc} fashion \revisedtext{\cite{Bellicoso2018a,mpc_oriolo,DiCarlo2018a}}. 
	%lack of descriptuiveness of reduced models
	The use of reduced models results in smaller optimization problems and 
	shorter computation times, at the price of a lower accuracy. 
	This is because reduced  models are often written in a reduced set of the state variables \revisedtext{and}
	capture the main dynamics of the robot during locomotion, but   
	typically neglect the joint dynamics. 
	Therefore, constraints at the joint variables (\eg torque or kinematic limits)
	cannot be explicitly formulated in the planning problem \revisedtext{(i.e. they lack descriptiveness)}.
	
	\begin{figure*}[h]
		\centering
		\includegraphics[width=1\textwidth]{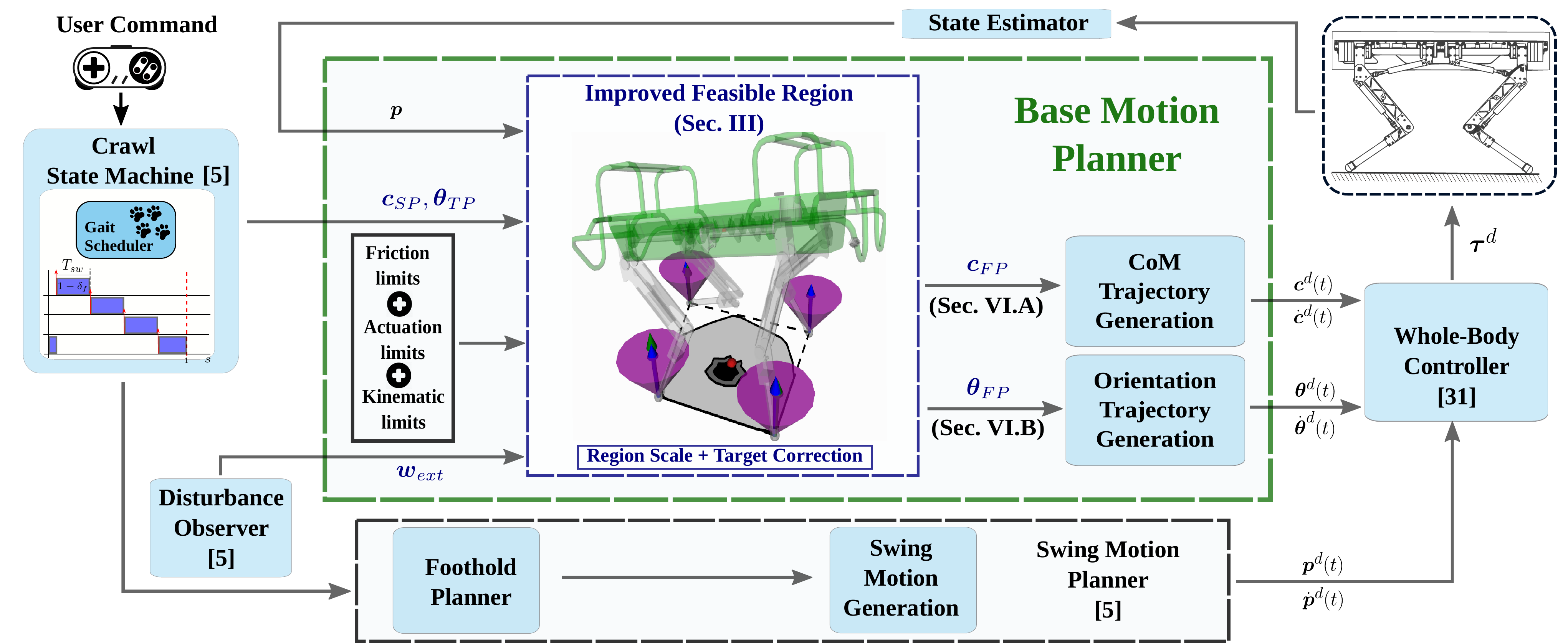}
		\caption{\small Block diagram of our locomotion framework. The improved feasible region is a help 
			for the planner to devise feasible robot postures.}
		\label{fig:crawl_diagram}
	\end{figure*}
	
	%idea of polytopes
	Borrowing ideas from computational geometry, researchers 
	succeeded in adding more descriptiveness to the 
	centroidal dynamics model without explicitly optimizing 
	for joint torques nor for  contact forces.
	This can be achieved by mapping  friction limits  (defined at the contact level) 
	and joint-torque limits  (defined at the joint level) 
	to the 6D space where the centroidal wrench exists. 
	%CWC and FWP
	These mappings result in 6D polytopes that represent 
	the set of admissible wrenches for which 
	the above-mentioned constraints are satisfied. 
	Namely, the \gls{cwc} is defined when only friction constraints 
	are considered \cite{Hirukawa2006, caron15}, while the \gls{fwp} is defined
	when both the friction and joint-torque limits are taken into account \cite{orsolino18ral}. 
	Enforcing the polytopes as constraints on the 
	centroidal wrench (or accelerations) in a \gls{to} problem
	results in feasible trajectories for
	the \gls{com}.
	%limits of FWP slow to compute
	Unfortunately, despite the promising results, the introduction of the 
	joint-torque limits  made the computation prohibitively expensive. In fact, increasing the number of contacts dramatically increases the computation time.
	This makes these polytopes hard to \revisedtext{compute} \textit{online} without accepting 
	strong approximations on kinematics \cite{orsolino18ral}.
	
	Another approach to address the problem of feasibility is to define 
	a reference point\footnote{In robotics there are many "ground" reference points
		used to devise locomotion strategies: ICP, ZMP, \gls{com}, etc. Here a reference point could be any generic point that is connected with the motion of the robot.}
	(henceforth we will consider the \gls{com}, 
	even though any other point can be chosen \cite{caron2016tro}) along with a 2D region
	in which the projection of the reference point must lie inside, in order to meet the
	requested feasibility  conditions (\eg friction, joint-torque or kinematic).
	
	Because of the above reasons, the feasible region 
	represents an intuitive yet powerful 
	way to plan feasible trajectories for the \gls{com} 
	while being favored with its computational efficiency. 
	%iperative projection
	Indeed, these regions are efficiently generated 
	through incremental projection algorithms \cite{kelley} that achieve a reduced computational complexity. 
	\revisedtext{Similar to the \gls{cwc} and \gls{fwp}, the feasible region could be enforced as constraints on the \gls{com} trajectory in a \gls{to} problem.}

	Bretl et al. \cite{Bretl2008a} were the first 
	to introduce an \gls{ip} algorithm for the computation of 
	a \textit{support region}
	for arbitrary terrain (\eg non coplanar contacts). We will refer to such region as the \textit{friction region} in the remainder of this manuscript to avoid possible confusion with the support polygon,
	which is the convex hull of the supporting feet.
	%
	
	%feasible region
	In our previous work \cite{orsolino19tro}, we \revisedtext{presented}
	a modified version of the IP algorithm to compute the \textit{feasible region}, 
	a convex region where both friction and joint-torque
	limits (\ie joint torque limits) were considered.
	As in the case of the \gls{fwp}, the \textit{feasible region} 
	varies with the contact condition and with the joint configuration.
	The advantage of this convex region 
	with respect to the 6D wrench polytope counterpart, is that it can be computed at least 20 times faster (15 $ms$ \revisedtext{on an i7-8700 3.2 GHz processor and 16GB of RAM}).
	This makes planning \gls{com} trajectories and foothold locations on 
	arbitrary terrains based on such region, suitable for \textit{online} implementation.
	
	%limitin assumptions
	Nonetheless, to simplify the analysis, a few assumptions were adopted \revisedtext{in \cite{orsolino19tro}} during the computation of the feasible region:
	($1$) the \revisedtext{only} external \revisedtextthird{wrench} acting on the
	robot is gravity,
	($2$) inertial accelerations and angular 
	dynamics are neglected (quasi-static assumption);
	this means that the model used to build the region is 
	a point mass model with contact forces,
	($3$) kinematic limits are not considered, and
	($4$) the region is always constructed on a plane perpendicular to gravity, making it not general enough
	to plan trajectories in  planes with different inclinations (\eg when climbing ramps).
	
	Because of assumption ($1$), the feasible region is incapable of capturing
	the effects of the application of an external wrench to the robot; external wrenches usually cause a shift in the region as well as a change in its shape and size (as will 
	be shown in section \ref{sec:wrenches}).
	Therefore, any planning strategy based on this region would be inaccurate and  can 
	lead to unfeasible plans when external disturbances are applied.
	Such a feature is also needed when an external \revisedtextthird{wrench} is \textit{intentionally} applied to the robot.
	%rope usage
	This is the case when  a load is pulled or when a rope is used for locomotion.
	In fact, having a feasibility metric that takes into account the 
	effect of external wrenches would open  many research opportunities 
	in rope-aided locomotion and load-pulling applications. External disturbances are incorporated in an \gls{mpc} in \cite{filippo} and \cite{ludovic} to plan stable \gls{zmp} trajectories. The method, however, utilizes the more simplified \gls{lip} model and is not suitable for non coplanar contacts.
	Furthermore, the restricting effect of the joint-torque limits on the CoM planning, in the presence of counteracting disturbances, is not considered.
	
	Assumption ($2$) limits the applicability of the feasible region to quasi-static gaits. 
	If applied to more dynamic gaits, having a trajectory computed 
	under a statically stable assumption 
	may induce falling due to the changes in the velocity of the robot.
	Recently, Audren et al. \cite{Audren2017} incorporated the dynamics,
	proposing a robust static stability region that accounts for possible \gls{com} accelerations. 
	No other feasibility measures such as joint torque and kinematic limits were considered.
	In contrast, Nozawa et al. \cite{cfr2016} compute a dynamic stability region for the \gls{com} based on specified linear and angular accelerations. In both approaches, however, only  friction guarantees were considered in the regions. 
	
	%kinematics 
	In addition, not accounting for kinematic limits in assumption ($3$) can be problematic 
	when the robot climbs up and down high obstacles or  
	is forced to walk in confined environments. 
	In such situations, the mandatory adjustments in \textit{height} 
	and \textit{orientation} may push the robot to violate its kinematic limits.
	In this respect, the seminal work of Carpentier et al. \cite{carpentier2017} 
	focused on incorporating the kinematic constraints via learning proxy constraints.
	On a similar line, \cite{tonneau2pac, Fernbach18}
	constrain the position of the \gls{com} with respect to the contact points,
	however, these kinematic constraints are only approximated to maintain the convexity of the problem,
	thus only valid for a simplified representation of the robot.
	More recently, Fankhauser et al. \cite{Fankhauser2018a} 
	optimized the orientation to ensure static stability and kinematic limits,
	by solving a non-linear optimization problem (SQP).
	The kinematic limits were roughly approximated by setting bounds on the leg length.
	An SQP problem is also utilized in \cite{cfr2016} to find a kinematically valid \gls{com} target
	close to the original target chosen solely on the stability region.  
	In the context of manipulators that  move assembly objects,
	other approaches \cite{mattikali, mosemann} 
	present a way to find all the  orientations that 
	satisfy static stability. 
	Yet, the objects were fixed and not actuated.

	\subsection{Contribution}
	In this work we aim to address the above limitations and 
	extend the descriptive capability of 2D admissible regions 
	by introducing a redefinition of the \textit{feasible region} initially proposed in \cite{orsolino19tro}. In particular we:
	
	\begin{itemize}[leftmargin=*]
		\item Generalize the feasible region to account for the effect of external wrenches  (see Section \ref{sec:wrenches}). Unlike in \cite{filippo} and \cite{ludovic}, the centroidal dynamics model is used and we consider external forces and torques acting on arbitrary points of the robot.
		\item We relax the quasi-static assumption by considering the dynamic effects, 
		as well as the angular dynamics \revisedtext{(see Section \ref{sec:dynamics})}.
		Differently from \cite{Audren2017} where the region 
		was built considering the set of admissible \gls{com} accelerations, 
		we consider the \textit{actual} acceleration resulting 
		in a time-varying shape of the region when the robot is in motion.
		\item Generalize the feasible region to be defined on arbitrary plane inclinations \revisedtext{(see Section \ref{sec:generic_plane})}. 
		\item Embed the complete joint-kinematic limits in what we define as the \textit{reachable region}. This presents a more accurate representation of the CoM kinematic capability than the approximations performed in \cite{tonneau2pac,Fernbach18, Fankhauser2018a}. Furthermore, the region does not need to be relearnt for different robots as in \cite{carpentier2017}. The region can be intersected with the joint-torque aware region \revisedtext{(with the aforementioned extensions)} and leads to the so-called \textit{improved feasible region} that considers friction, joint-torque and kinematic limits.
		\item \revisedtext{Design a \textit{robust} \gls{com} planning strategy that utilizes non-convex regions} and propose a new optimization for the trunk orientation based solely on the improved feasible region.
		The level of robustness can be adjusted 
		by tuning a single parameter according to the 
		desired level of "cautiousness" one wants to achieve in the locomotion. 
		\item Show simulations and hardware experiments with robots walking in scenarios that are
		challenging in terms of actuation and kinematic motions. We compare 
		a planning approach based on the \textit{improved feasible region} 
		with our previous heuristic approach \cite{Focchi2020}.
		The experimental results are shown on both the 90 kg HyQ robot (hydraulically actuated) and the 21 kg Aliengo (electrically actuated) robot \cite{aliengo}. Feasibility of dynamic motions are validated by the feasible region in experiments with the Aliengo robot.
	\end{itemize}

	\subsection{Outline}
	The paper is organized as follows: in Section \ref{sec:recap_feasible}
	we recall the modified \gls{ip} algorithm used to compute 
	the feasible region \revisedtext{\cite{orsolino19tro}}, while in Section \ref{sec:ext_feasible} we present the extensions to compute the feasible region. Section \ref{sec:kin_lim} introduces the \textit{reachable region} and Section \ref{sec:improved} defines the intersection of the \textit{feasible region} and the \textit{reachable region} to define the \textit{improved feasible region}.
	Section \ref{sec:planning} illustrates the planning strategies for the \gls{com}
	and the orientation based on the region.
	\revisedtextthird{Section \ref{sec:assumptions} summarizes the assumption made in this work. }
	Simulations and experimental results with HyQ \revisedtext{and Aliengo robots} are presented in Section \ref{sec:simulations} and \ref{sec:experiments}.
	Section  \ref{sec:conclusions} draws the conclusions and discusses
	possible future developments.
	Finally, the Appendix includes additional information about the reachable region and its effect on planning.
	
	\section{Recap on classical feasible region}
	\label{sec:recap_feasible}
	
	%Review of the usual algorithm, recap on equations
	\revisedtext{For a better understanding of the proposed improved feasible region, let us first briefly recap the \textit{feasible region} presented in \cite{orsolino19tro}.
		This region was generated using an \gls{ip} algorithm described in Algorithm \ref{alg:IP} (in black). The extension of the region to generate the improved feasible region are marked in \textcolor{dukeblue}{blue} and described in detail in Section \ref{sec:ext_feasible}.}
	
	The algorithm considers the convex constraints imposed on a legged robot and projects them onto a 2D linear subspace.
	This is done by building an inner and outer approximation of the projected region, 
	via iteratively solving a sequence of LP programs while satisfying the convex constraints 
	(shown in step (III) of Algorithm \ref{alg:IP}).
	Namely, we considered  the static stability constraints (III.a), 
	frictional constraints on the contact feet (III.b), and  the joint-torque constraints (III.c).
	
\begin{figure}[tb]
		\centering
		\includegraphics[width=0.35\textwidth]{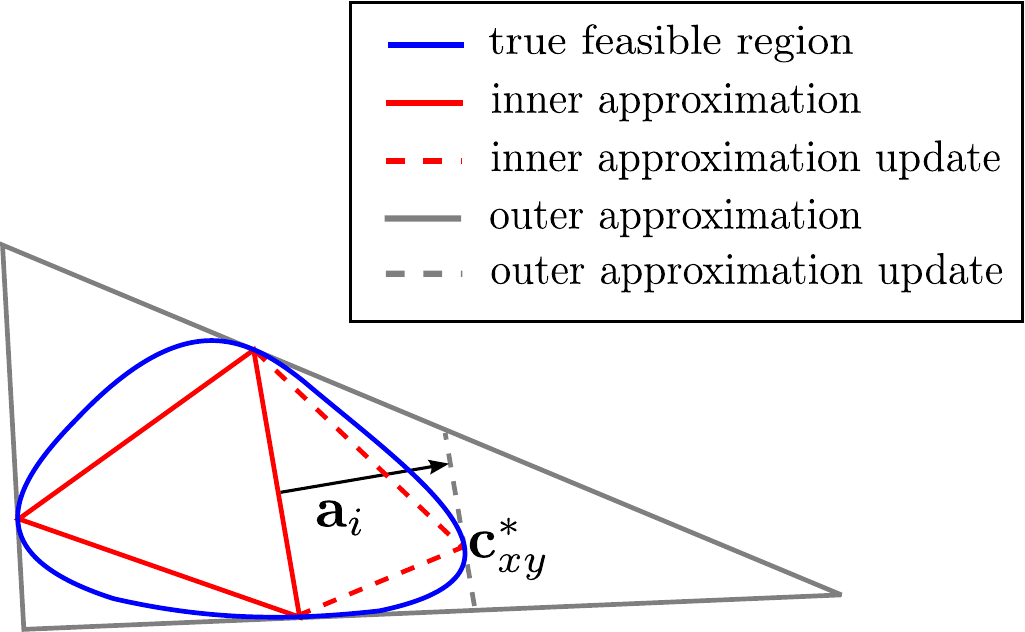}
		\caption{\small Iteration of the \gls{ip} algorithm: after the LP is solved finding a new extremal $\mathbf{c}^*_{xy}$ point along $\mathbf{a}_i$, this is added to the \textit{inner} approximation while an edge with normal $\mathbf{a}_i$ passing through $\mathbf{c}^*_{xy}$ is added to the  \textit{outer} approximation \cite{orsolino19tro}.}
		\label{fig:ip_update}
	\end{figure}

	The solution of each LP problem, $\mathbf{c}_{xy}^*$, 
	is an extremal CoM position along a certain direction (represented by the unit vector $\mathbf{a}_i$),
	that still satisfies the constraints, \ie a vertex on the boundary of the feasible region.
	This optimization is performed iteratively along various directions $\mathbf{a}_i$ that span along a circle, building the inner approximation of the region as the convex hull of all the solutions $\mathbf{c}_{xy}^*$ (see Fig. \ref{fig:ip_update}).

	\begin{algorithm}
		\begin{algorithmic}
			\State \textbf{Input: }{$\mathbf{c}_{xy}, \textcolor{dukeblue}{c_z}, 
				{}^\mathcal{W}R_\mathcal{B},  \mathbf{p}_1, ..., \mathbf{p}_{n_c}, \mathbf{n}_1, ..., \mathbf{n}_{n_c}, \mu_1, ..., \mu_{n_c},$
				\State \hspace{1.0cm} $\boldsymbol{\underline\tau}_1, ..., \boldsymbol{\underline{\tau}}_{n_c}, \boldsymbol{\bar\tau}_1, ..., \boldsymbol{\bar\tau}_{n_c}, \textcolor{dukeblue}{\boldsymbol{w}_{ext}}$}
			\State \textbf{Result: } local feasible region $\mathcal{Y}_{fa}$
			\State \textbf{Initialization}: $\mathcal{Y}_{outer}$ and $\mathcal{Y}_{inner}$\;
			\While{$area(\mathcal{Y}_{outer}) - area(\mathcal{Y}_{inner}) > \epsilon$}
			\State I) compute the edges of $\mathcal{Y}_{inner}$\;
			\State II) pick $\mathbf{a}_i$ based on the edge cutting off the largest fraction of $\mathcal{Y}_{outer}$\;
			\State III) solve the LP:\:
			
			\vspace{0.2cm}${\begin{aligned}
					\mathbf{c}_{xy}^* = \argmax_{\mathbf{c}_{xy}, \mathbf{f}} \quad &  \mathbf{a}_i^T \mathbf{c}_{xy}\\
					\text{such that}:\\
					(\text{III.a}) &	\quad
					\mathbf{A}_1 \mathbf{f} + \textcolor{dukeblue}{\mathbf{A}_2} \mathbf{c}_{xy} = \textcolor{dukeblue}{\mathbf{u}}\\
					(\text{III.b}) &	\quad    \mathbf{B}\mathbf{f} \leq \mathbf{0} \\
					(\text{III.c}) &	\quad    \mathbf{G}\mathbf{f} \leq \mathbf{d} 
			\end{aligned}}$
			%			\State \:
			\State IV) update the outer approximation $\mathcal{Y}_{outer}$\;
			\State V) update the inner approximation $\mathcal{Y}_{inner}$\;
			\EndWhile
			
		\end{algorithmic}
		\caption{Feasible Region IP algorithm \textcolor{dukeblue}{(with external wrenches)}.}
		\label{alg:IP}
	\end{algorithm}
	
	Constraint (III.a) ensures the static balance of the robot (force and moment balance).
	%All vectors, unless specified, are expressed in an inertial (world) frame.
	$\mathbf{A_1} \in \Rnum^{6\times m {n_c}}$ is the grasp matrix of 
	the  $n_c$ contact points  $\mathbf{p}_i \in \Rnum^3$  and $m$ depends on the nature of the contact (\ie $m=3$ for point contact, $m=6$ for full contact).
	$\mathbf{A_1}$ is summing up the contact wrenches (pure forces in case of point feet) $\mathbf{f} \in \mathbb{R}^{mn_c}$ 
	and is expressing them at the origin of the world frame.
	$\mathbf{u} \in \mathbb{R}^{6}$ is the linear part of the wrench due to gravity force (acting on the \gls{com}) 
	and $\mathbf{A_2}$ computes the angular component of the gravity wrench, whenever this is
	expressed at the origin of the world frame:
	%%%%%%%%%%%%%
	\begin{equation}
		\begin{aligned}
			&\mathbf{A}_1 = \mat{\bar{\mathbf{A}}_1 & \dots & \bar{\mathbf{A}}_{n_c}} \in \Rnum^{6\times m {n_c}}, \\
			& 	\mathbf{A}_2 = \mat{\mathbf{0} \\
				-m \mathbf{g} \times \mathbf{P}_{xy}^T} \in \Rnum^{6\times 2}, \quad \mathbf{P}_{xy} = \mat{1 & 0 & 0\\
				0 & 1 & 0}\\
			&	\mathbf{u} = \mat{ - m \mathbf{g} \\ \mathbf{0}},
			\hspace{2.7cm} \mathbf{g} = [0,0,-g]^T.
		\end{aligned}
	\end{equation}
	$\mathbf{P}_{xy}$ is the selection matrix selecting the 
	horizontal components $x,y$ of the \gls{com} and $\mathbf{\bar{A}}_i$ is such that:
	\begin{equation*}
		\bar{\mathbf{A}}_i = \left\{ \begin{tabular}{ll}
			\vspace{0.1cm}
			\mat{\mathbf{I}_3 \\
				[\mathbf{p}_i]_\times} $ \in \Rnum^{6\times 3}$ & \text{if} \quad $m = 3$\\ 
			\mat{\mathbf{I}_3 & \mathbf{0}_3\\
				[\mathbf{p}_i]_\times & \mathbf{I}_3}  $\in \Rnum^{6\times 6}$ & \text{if} \quad $m = 6$ \\
		\end{tabular} \right.
	\end{equation*}
	where $[\cdot]_\times$ is the skew-symmetric matrix associated to the cross product.
	
\begin{figure}[tb]
		\centering
		\includegraphics[width=0.7\linewidth]{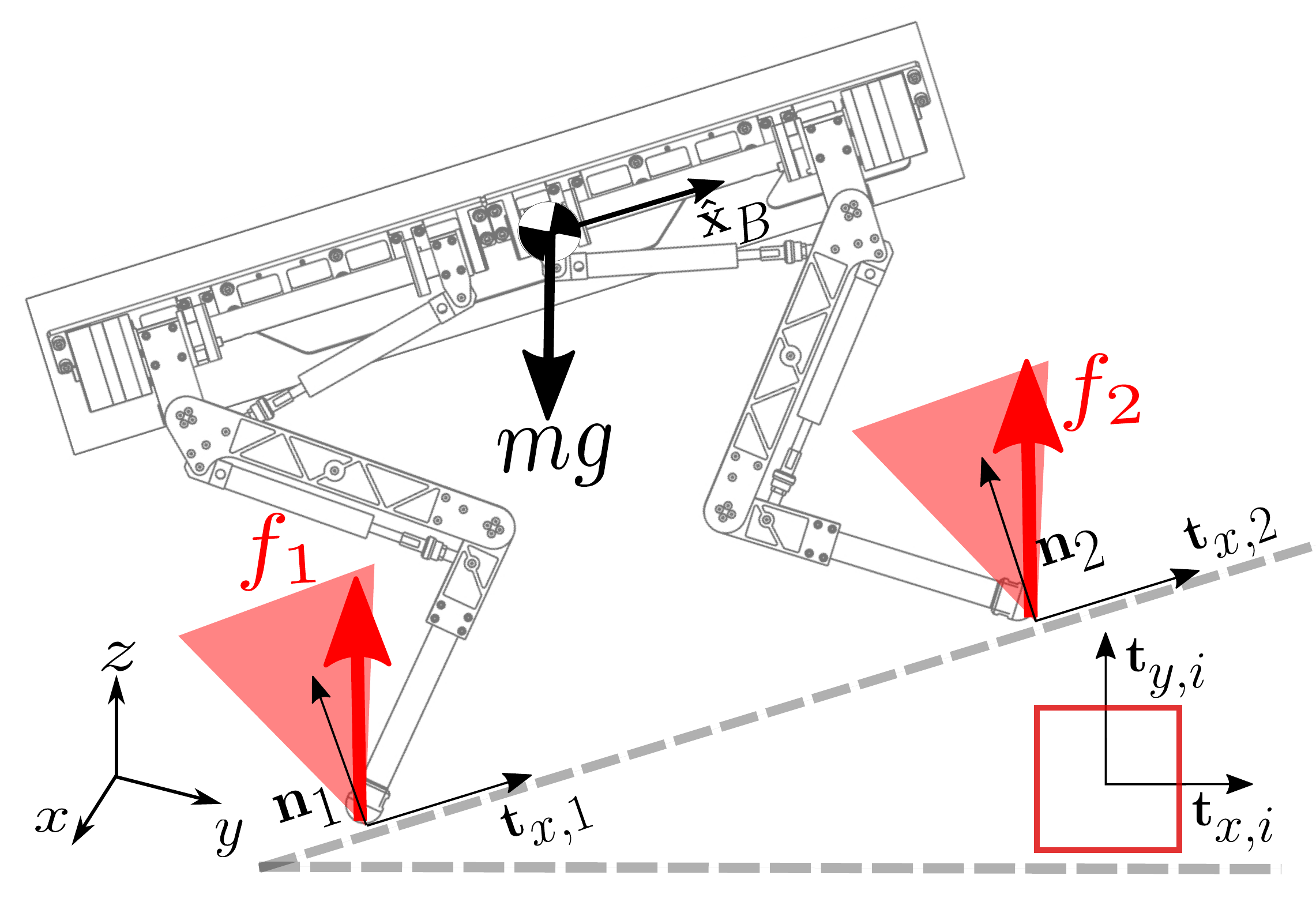}
		\caption{\small \revisedtextthird{Friction pyramids (shown in red) orientation with respect to the contact surface. Each pyramid base is perpendicular to the contact surface normal $\mathbf{\hat{n}}_i$. A top view of the pyramid base with respect to the tangent contact axes $\mathbf{\hat{t}}_{x,i}, \mathbf{\hat{t}}_{y,i}$ is shown (bottom right).}}
		\label{fig:ramp_pyramids}
	\end{figure}

	Constraint (III.b) ensures the friction constraints are met.
	These require the contact forces to be inside \revisedtext{inner pyramidal} (conservative) approximations of the  friction  cones. Approximating the friction cones with a low number of linear approximations results in a smaller computation time \cite{friction1,friction2,frictionhandbook}. The number of edges chosen to represent the pyramid with reasonable accuracy can be chosen based on the complexity of the terrain and the friction coefficient. \revisedtextthird{For each contact, we can define an orthonormal reference frame composed of the contact surface normal $\mathbf{\hat{n}}_i \in \Rnum^3$, and tangent vectors $\mathbf{\hat{t}}_{x,i}, \mathbf{\hat{t}}_{y,i} \in \Rnum^3$ such that $\mathbf{\hat{t}}_{x,i} = \mathbf{\hat{n}}_i \times \mathbf{\hat{x}}_B \times \mathbf{\hat{n}}_i$, where $\mathbf{\hat{x}}_B$ is the unit vector along the X-axis of the base of the robot. Each pyramid is oriented along $\mathbf{\hat{n}}_i$ (with the base of the pyramid parallel to the contact surface) as shown in Fig. \ref{fig:ramp_pyramids}.
	T}he constraint matrix $\mathbf{B} \in \mathbb{R}^{4n_c \times 3n_c}$ \revisedtextthird{can then be} represented as:
	\begin{equation}\label{eq:force_constraint_matrix}
		\begin{aligned}
			\mathbf{B} &= \diag(\mathbf{b}_1, \dots, \mathbf{b}_{n_c}),\\
			\mathbf{b}_i &= \mat{
				(\mathbf{\hat{t}}_{x,i} - \mu_i \mathbf{\hat{n}}_i)^T\\ (\mathbf{t}_{y,i} - \mu_i \mathbf{\hat{n}}_i)^T \\ -(\mathbf{\hat{t}}_{x,i} + \mu_i \mathbf{\hat{n}}_i)^T \\ - (\mathbf{\hat{t}}_{y,i} + \mu_i \mathbf{\hat{n}}_i)^T} \in \Rnum^{4 \times 3} \\
		\end{aligned}
	\end{equation}
	%
	%bilateral constraints
	%
	Finally, constraint (III.c) ensures that the  torque at each joint does not exceed its limit.
	These limits are mapped to the end-effector (feet) 
	space by means of the inverse-transpose of the Jacobian\footnote{This is true for a non-redundant leg, where the Jacobian is a square matrix.}.
	This yields to the definition of force polytopes that represent the sets of 
	admissible contact forces that respect joint-torque limits.
	By considering   the vectors of minimum  ($\boldsymbol{\underline\tau}_i \in \mathbb{R}^{n_l}$) 
	and maximum  ($\boldsymbol{\bar\tau}_i \in \mathbb{R}^{n_l}$) joint torque limits,  on the $n_l$ joints of  the $i$th leg,
	the half-plane description of such force polytopes is represented by $\mathbf{G} \in 
	\mathbb{R}^{2 n_l n_c \times m n_c}$ and $\mathbf{d} \in \mathbb{R}^{2 n_l n_c}$:
	\begin{equation}
		\begin{aligned}
			\mathbf{G} &= \text{diag}\left(   \mat{\mathbf{J}(\mathbf{q}_1)^T \\-\mathbf{J}(\mathbf{q}_1)^T}, \dots, \mat{\mathbf{J}(\mathbf{q}_{n_c})^T \\ - \mathbf{J}(\mathbf{q}_{n_c})^T} \right), \mathbf{d} = \mat{ \mathbf{d}_1 \\\vdots \\ \mathbf{d}_{n_c}} 
		\end{aligned}
	\end{equation}
	where $\mathbf{q}_i$ represents the vector of angular 
	positions of the joints of the $i$-th leg in contact with the environment 
	(cfg.  \cite{orsolino18ral} on how to compute  $\mathbf{d}$  
	from $\boldsymbol{\underline\tau}$ and $\boldsymbol{\bar\tau}$).
	Because $\mathbf{G}$ and $\mathbf{d}$ are configuration-dependent, the force polytopes and the 
	resulting feasible region are, thus, only locally valid in a neighbourhood of the considered instantaneous configuration.
	\revisedtext{The consequence of this is that the feasible region can be considered to be accurate only
		in a neighborhood of the considered robot configuration.}
	Therefore, for every change in the  \gls{com} position
	due to a change in the joint configuration, 
	the feasible region should be recomputed.
	
	With this, we can formally define the \textit{feasible region} 
	encompassing all the \gls{com} positions $\mathbf{c}_{xy}$ that satisfy 
	the friction constraints and the joint-torque constraints simultaneously as:
	\begin{equation}
		\mathcal{Y}_{fa} = \Big\{\mathbf{c}_{x y} \in \mathbb{R}^{2} | \quad  \exists \mathbf{f}_i \in \mathbb{R}^{m n_c}, \text { s.t. } (\mathbf{c}_{x y}, \mathbf{f}_i) \in \mathcal{C} \cap \mathcal{A}\Big\}
	\end{equation}
	where $\mathcal{C} \cap \mathcal{A}$ is the set of contact forces and \gls{com}  positions 
	(projected on an $X-Y$ plane) satisfying both friction and joint-torque constraints:
	\begin{equation}
		\begin{aligned}
			\mathcal{C} \cap \mathcal{A} = \Big\{\mathbf{f}_i \in \mathbb{R}^{m n_c}, \mathbf{c}_{x y} \in \mathbb{R}^{2}& | \quad  \mathbf{A}_1 \mathbf{f} + \mathbf{A}_2 \mathbf{c}_{xy} = \mathbf{u}\\ &\mathbf{B}\mathbf{f} \leq \mathbf{0}, \quad \mathbf{G}\mathbf{f} \leq \mathbf{d} \Big\}
		\end{aligned}
	\end{equation}

	\section{Feasible Region Extensions} 
	\label{sec:ext_feasible}
	In this section we propose an extension of the feasible 
	region to arbitrary
	plane inclinations (Section \ref{sec:generic_plane}). We then proceed to incorporate external wrenches (Section \ref{sec:wrenches}), and
	dynamic effects (Section \ref{sec:dynamics}).
	The changes on the algorithm are highlighted in \textcolor{dukeblue}{blue} in Algorithm \ref{alg:IP}.
	
	\subsection{Generic Plane of Projection}\label{sec:generic_plane}
	%extesion to a generic plane
	Under the sole influence of gravity and considering only \textit{friction} constraints,
	the static equilibrium constraints in \cite{Bretl2008a} 
	are only affected by the horizontal position of the \gls{com}\footnote{The only dependence on the CoM position 
		is due to $\mathbf{c} \times m \mathbf{g} = m||g|| \begin{bmatrix} 
			-c_y & c_x & 0
		\end{bmatrix}^T$ in the moment balance constraints. The zero in the last row shows the independence from the vertical coordinate of the CoM.
	}. 
	Therefore, the high dimensional constraints were naturally projected 
	on a plane perpendicular to gravity (\ie the horizontal plane).
	In such case, for a given set of contacts, checking stability for a \gls{com} 
	trajectory with a varying height is still appropriate with respect to the projected region.
	However, when used for planning purposes, computing the region in a plane consistent with the 
	planned motion can be of convenience.
	One would then simply need to find a feasible 2D \gls{com} 
	trajectory in the plane of reference.
	Therefore it is important to have the possibility to choose the plane of interest where the region is computed.
	
	More importantly, as will be explained further in Section \ref{sec:wrenches}, 
	under the influence of external and inertial wrenches on the \gls{com}
	(and when including joint torque and kinematic constraints), the \gls{com} vertical position 
	can alter the feasible region.
	Therefore,  for a given set of contacts,  the feasible region
	will  be dependent on the height of the robot;
	in this case, planning a \gls{com} motion defined in 
	a plane \textit{inconsistent} with the one used for the 
	computation of the region, could result in infeasibility.
	Thus, to compute the region, it is important to project the high dimensional constraints 
	on the plane where the expected \gls{com} trajectory will lie.
	
	For instance, for a robot climbing a ramp, the planned \gls{com} trajectory can be expected to 
	follow the inclination of the ramp \cite{Focchi2020}\cite{gehring2015}. In general, the orientation of the projection plane depends on the planning strategy: choosing a plane of projection consistent with the terrain inclination and with the \gls{com} trajectory ensures a constant \gls{com} height when expressed with respect to such plane.\footnote{Note that the projection the \gls{ip} algorithm is in fact a mapping of the high dimensional constraints from the
		wrench space (or in the case of the kinematic constraints, the joint space) to a Euclidean plane. The Euclidean plane can be chosen to be expressed with respect to the frame of our choice as explained above.}
	
	The inclination of a generic plane of interest $\Pi$ can be 
	described through a free vector $\mathbf{\hat{n}}$ 
	normal to it (expressed with respect to the world frame).
	Constraints (III) can be projected on to the plane of interest $\Pi$ 
	by applying the following change of coordinates:
	\begin{equation}
		\mathbf{c} =  {}^\mathcal{W}\mathbf{R}_\Pi \mathbf{\hat c}
		\label{eq:transformation}
	\end{equation}
	where $\mathbf{c} = [\mathbf{c}_{xy}^T \ c_z]^T$ and $\mathbf{\hat{c}} = [\mathbf{\hat{c}}_{\hat{x}\hat{y}}^T \ \hat{c}_{\hat{z}}]^T$
	are the \gls{com} position expressed with respect to the world 
	frame $\mathcal{W}$ and a frame attached 
	to the plane of interest $\Pi$, respectively.
	${}^\mathcal{W}\mathbf{R}_\Pi$ is the rotation matrix representing 
	the orientation of the plane of interest $\Pi$ with respect to the world frame $\mathcal{W}$, and is defined as:
	\begin{equation}
		{}^\mathcal{W}\mathbf{R}_\Pi = \mat{\mathbf{\hat{x}}, \mathbf{\hat{y}},	\mathbf{\hat{z}} }
	\end{equation}
	The $\hat{z}$-axis of $\Pi$ is aligned with $\mathbf{\hat
	{n}}$.
	$\mathbf{\hat{x}}, \mathbf{\hat{y}}$ are unit vectors 
	(expressed in $\mathcal{W}$ frame and forming the $\hat{x},\hat{y}$-axes of $\Pi$ frame)
	chosen such that they form, together with $\mathbf{\hat{z}}$, a right-handed coordinate system.
	With the change of coordinates in \eqref{eq:transformation},
	the IP algorithm can be written in terms of $(\mathbf{\hat c}_{\hat{x}\hat{y}}, \hat c_{\hat{z}})$ and solved for the new coordinates $\mathbf{\hat c}_{\hat{x}\hat{y}}$.
	In the remainder of this manuscript, not to overload the notation, we express the \gls{com} position in the world frame $\mathbf{c}$ in all related equations, without any loss of generality.
	
	\subsection{External wrenches}
	\label{sec:wrenches}
	Consider an external wrench, $\boldsymbol{w}_{ext} = [\boldsymbol{f}_{ext}, \boldsymbol{\tau}_{ext}]^T 
	\in \mathbb{R}^6$, applied on the CoM of a legged robot.
	For the robot to be in \textit{static} equilibrium, 
	the wrench balance equations should satisfy:
	\begin{align}
		&	\sum_{i=1}^{n_c} \mathbf{f}_i + m\mathbf{g}  + \boldsymbol{f}_{ext} = 0\\
		&	\sum_{i=1}^{n_c} \mathbf{p}_i \times \mathbf{f}_i
		- (m \mathbf{g} + \boldsymbol{f}_{ext}) \times \mathbf{c}
		+ \boldsymbol{\tau}_{ext}= 0
		\label{eq:moment_balance}
	\end{align}
	As mentioned in the previous section, with only the gravity $\mathbf{g}$ acting on the robot,
	the dependence on the CoM in \revisedtext{(\ref{eq:moment_balance})} only comes from its horizontal positions $\mathbf{c}_{xy}$.
	However, with the presence of an external force, $\boldsymbol{f}_{ext}$,
	a dependence on the CoM vertical position $c_{z}$ can clearly exist
	from the term  $-\boldsymbol{f}_{ext} \times \mathbf{c}$ (unless $\boldsymbol{f}_{ext}$ is aligned with gravity).
	
	To incorporate the effect of $\boldsymbol{w}_{ext}$ on Algorithm \ref{alg:IP}, 
	the constraint (III.a) can be rewritten by redefining $\mathbf{A}_2$ and $\mathbf{u}$ to be:
	\begin{equation}
		\begin{aligned}
			\textcolor{dukeblue}{\mathbf{A}_2} &= 
			\begin{bmatrix}
				\mathbf{0} \\
				-[m \mathbf{g} + \boldsymbol{f}_{ext}] \times \mathbf{P}_{xy}^T
			\end{bmatrix} \in \Rnum^{6\times 2} \\
			\textcolor{dukeblue}{\mathbf{u}} &= 
			\begin{bmatrix}
				- m \mathbf{g} - \boldsymbol{f}_{ext}\\
				[\boldsymbol{f}_{ext}] \times \mathbf{P}_{z}^T c_z
				- \boldsymbol{\tau}_{ext}
			\end{bmatrix} \in \Rnum^{6\times 1}
		\end{aligned}
	\end{equation}
	Therefore, $\mathbf{A}_2$ computes the moments due to gravity and 
	external forces (acting on the robot CoM\footnote{If a pure force is applied 
		in a different point of the robot the equivalent wrench at \gls{com}  should be computed. }), about the origin of the world frame.
	
	To better appreciate the effect of an external wrench $\boldsymbol{w}_{ext}$ 
	on the projected region we can further inspect its direct influence on $\mathbf{c}_{xy}$.
	$\mathbf{c}_{xy}$ characterizes the set of all the projected feasible \gls{com} positions, 
	given the existence of feasible contact forces $\mathbf{f}$.
	From the first two equations in (\ref{eq:moment_balance}), $\mathbf{c}_{xy}$ can be determined as \revisedtextthird{\cite{caron2016tro}}:
	\begin{equation}
		\begin{split}
			\revisedtextthird{\mathbf{c}_{xy}} &
			\revisedtextthird{= \frac{1}{-mg + f_{ext,z}}
				\biggl(
				\begin{bmatrix}
					0&0&1
				\end{bmatrix}^T \times \sum_{i=1}^{n_c} \mathbf{p}_i \times \mathbf{f}_i
				- c_z \mathbf{f}_{ext,xy} }\\
			%				\begin{bmatrix}
				%				F_x & F_y
				%				\end{bmatrix}^T
			&\revisedtextthird{+ \begin{bmatrix}
					-\tau_{ext,y} & \tau_{ext,x}
				\end{bmatrix}^T
				\biggr)}\\ 
			&= -\mathbf{h}(\mathbf{f},\revisedtextthird{f_{ext,z}}) + \mathbf{m}(\boldsymbol{f}_{ext}, \boldsymbol{\tau}_{ext}, c_z)
		\end{split}
	\end{equation}
	From \revisedtextthird{the offset function $\mathbf{m}$}, one could observe that an external wrench 
	applied on the robot, combined with the \gls{com} vertical position,
	results in a shift in the location of the projected \gls{com} positions (\ie projected region).
	
	The change in shape of the region, can be intuitively understood,
	considering that the set of contact forces resulting from the action 
	of the external wrench, could  become infeasible due to the additional 
	effort needed to compensate for the external wrench.	
	For example, in case of a significantly retracted leg, because the 
	joint-torques are propagated through the leg to the foot via the Jacobian, 
	the \gls{com} positions closer to the contact feet are more likely to be infeasible. 
	Furthermore, a \gls{com} projection located near a specific foot,
	further loads that foot (while reducing the load on the other feet).
	This drives the joints of that leg closer to their 
	torque limits making this \gls{com} position more likely to be infeasible.
	This explains why an external wrench applied on the robot, such as an additional load, results in smaller feasible regions as opposed 
	to the case when only the weight of the robot has to be supported.
	
\begin{figure}[tb]
		\centering
		\begin{subfigure}{8cm}
			\includegraphics[width=1\textwidth]{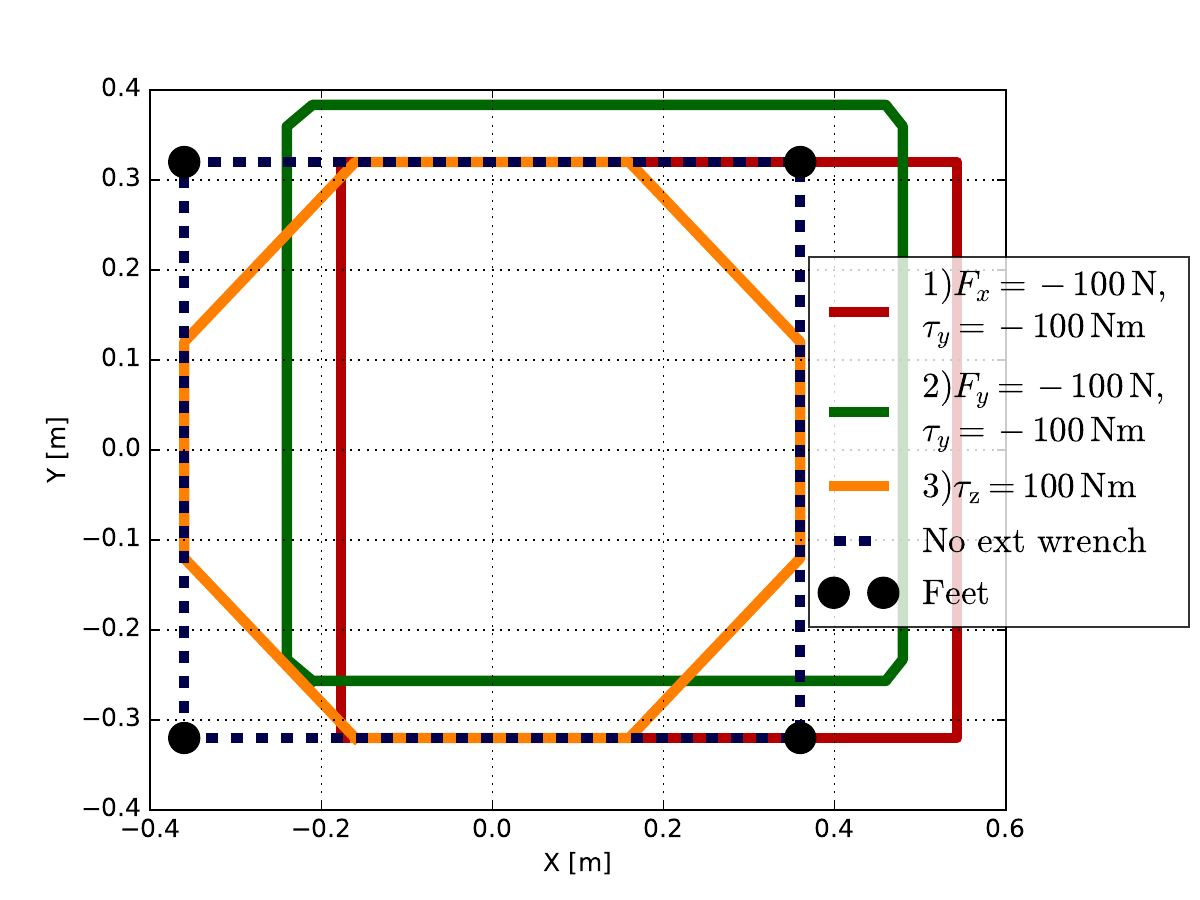}
			\caption{Friction Region (only friction considered)}
		\end{subfigure}\\
		\begin{subfigure}{8cm}
			\includegraphics[width=1\textwidth]{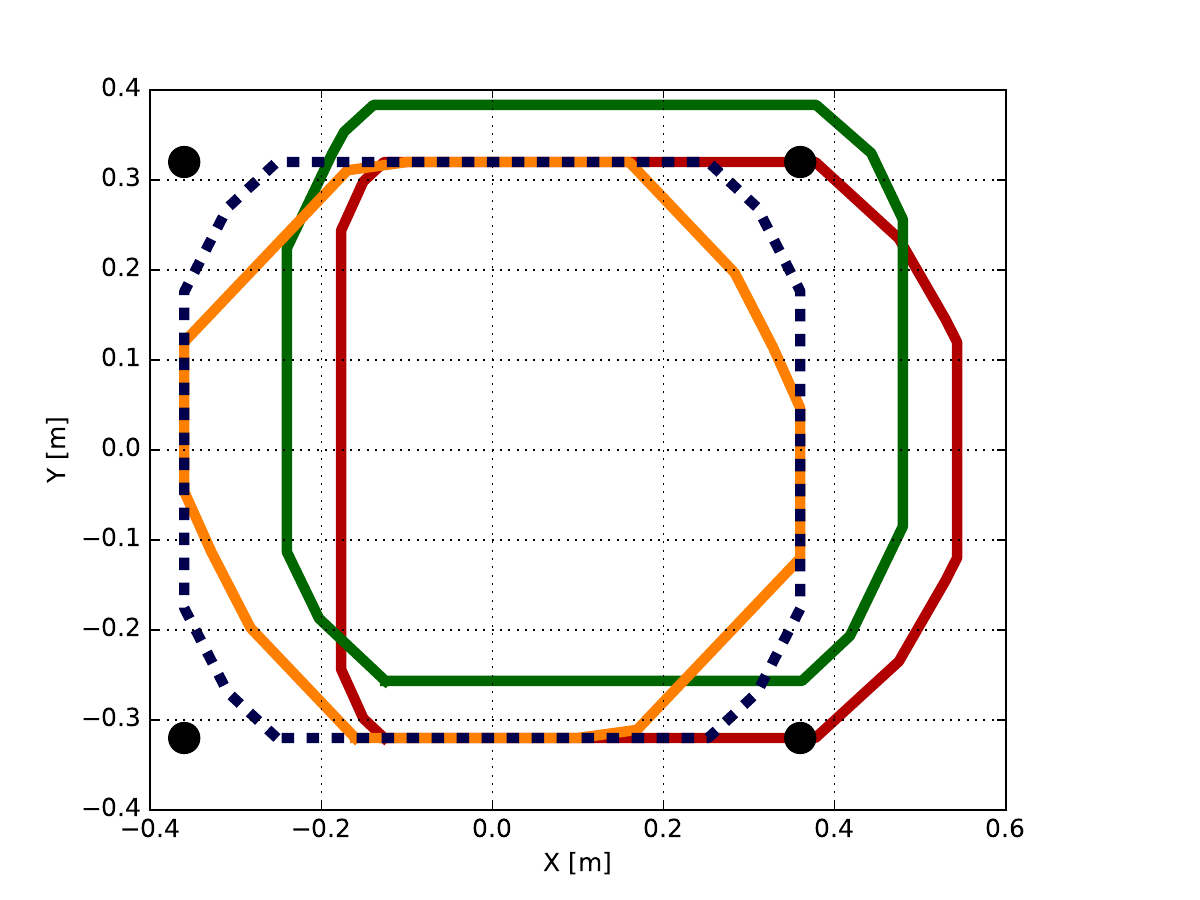}
			\caption{Feasible Region (both friction and joint-torque limits considered)}
		\end{subfigure}
		\caption{\small Effect  of different external wrenches acting on the CoM on the (a) friction region and the (b) feasible region.
			Changes in size and shifting of the location of the regions can be observed. 
			The components of the external wrench that are \revisedtextthird{mentioned are applied simultaneously and the} \revisedtextthird{un}mentioned \revisedtextthird{components} are \revisedtextthird{set to} zero.
			The stance feet of \gls{hyq} are shown as black points with the front feet facing right. Regions are computed for a trunk height of $c_z = 0.53 m$}
		\label{fig:ext_wrench}
	\end{figure}

	Figure \ref{fig:ext_wrench} illustrates examples of the resulting
	friction and feasible regions for different external 
	wrench cases calculated for the \gls{hyq} robot at $c_z = 0.53 m$.
	Case 1 (red) and 2 (green) show a shift both in the friction and in the
	feasible regions in the opposite direction to the external \revisedtextthird{wrench}.
	A reduction in the size of the friction region (\eg obtained
	only considering friction constraints (III.b)) can also 
	be seen for an external torque $\tau_{ext,z}$ (orange).
	This is illustrated by the clipping of the corners of the region, 
	where no admissible set of contact forces 
	could withstand such external wrench without slipping.
	
	\subsection{Dynamic Motions} 
	\label{sec:dynamics}
	
	To ensure stability/feasibility, it is necessary that the chosen 
	reference point remains inside the admissible region that was computed for it. 
	To evaluate dynamic stability, it is common to consider the \gls{zmp} as specified 
	reference point. Because the \gls{zmp} already explicitly considers the horizontal
	acceleration of the robot's body, this does not have 
	to be considered in the computation of the admissible region: this region therefore 
	can be obtained for dynamic conditions and, on flat terrains, (if only friction cone 
	constraints are considered) it simplifies to the convex hull of the contact points. 
	Therefore, we underline that the choice of a reference point and its admissible region are 
	tightly coupled and that any arbitrary reference point could be used provided that 
	the employed admissible region is specifically formulated in accordance to it. 
	As long as this point is inside the corresponding computed region, we are sure that the constraints 
	that have been considered when building the region, are satisfied. 
	Therefore, conforming to the previous sections, we keep using the \gls{com} as the reference point and proceed to incorporate the dynamic effects (dropping the static assumptions) in the feasible region (constraints III.a in Algorithm \ref{alg:IP}).
	In fact, it could even happen that the \gls{zmp} is outside of the 
	computed region,
	yet dynamic stability is ensured and the robot configuration 
	is feasible as long as the \gls{com} projection is inside it.
	
	Note that, including dynamic effects requires that we express the Newton-Euler equations 
	in the inertial frame. This means that the moment balance 
	should be done with respect to the origin of the inertial frame,
	that in general is not coincident with the \gls{com}.
	\revisedtext{Then the expression of Newton-Euler equations becomes:
		\begin{equation}
			\begin{cases} 
				m \left( \mathbf{\ddot{c}} - \mathbf{g}\right)  = \sum_{i=1}^{n_c} \mathbf{f}_i \\
				\mathbf{I_G} \boldsymbol{\dot{\omega}} + \boldsymbol{\omega} \times \mathbf{I_G} \boldsymbol{\omega} + \mathbf{c} \times m \left( \mathbf{\ddot{c}} 	- \mathbf{g}\right) = \sum_{i=1}^{n_c} \mathbf{p}_i \times \mathbf{f}_i\\
			\end{cases}
			\label{eq:dynamics}
		\end{equation}
		where  $I_G \in \Rnum ^{3 \times 3}$ is the moment of inertia about the center of mass, 
		$\mathbf{\ddot{c}}$ the \gls{com} 
		Euclidean acceleration, and $\boldsymbol{\dot{\omega}},\boldsymbol{\omega}$ the 
		angular acceleration and velocity of the robot base, respectively.
		By inspecting \eqref{eq:dynamics} one can see that to incorporate the dynamic effects, 
		the matrix $\mathbf{A}_1$ remains unchanged while $\mathbf{A}_2$ and $\mathbf{u}$ in constraint (III.a) should be redefined as:}
		\begin{equation}
		\begin{aligned}
			&	\mathbf{A}_2 = \mat{\mathbf{0} \\
				-m \left(\mathbf{g}-\mathbf{\ddot{c}}\right) \times \mathbf{P}_{xy}^T}
			&	\mathbf{u} = \mat{  m \left(\mathbf{\ddot{c}} - \mathbf{g}\right) \\ \mathbf{I_G} \boldsymbol{\dot{\omega}} + \boldsymbol{\omega} \times \mathbf{I_G} \boldsymbol{\omega}},
			\label{eq:dynamics2}
			\setlength{\belowdisplayskip}{0pt}
		\end{aligned}
		\end{equation}
Note that now the \textit{simple mass} model becomes a \textit{centroidal dynamics} model 
as the angular dynamics is also taken into account. 
Moreover, the \textit{static} stability enforced in \revisedtext{constraint} (III.a) can be considered to be fully \textit{dynamic}.
As a result of the effect of the inertial accelerations, the computed region can "move" (\eg forward or backward) according to  the direction of the 
instantaneous body acceleration (see accompanying video \revisedtext{at 0:20 and 04:52}).
With the dependence of the feasible region on the acceleration of the robot, one can utilize the desired body accelerations in the computation of the region to plan dynamically feasible motions.
\subsection{Degenerate Feasible Regions}
\label{sec:degenerate}
It is possible to further extend the feasible region to  
dynamic gaits in quadrupeds (\eg a trot or pace) were only one or two point contacts are established with the ground at the same time. 
In these cases, the classical support polygon collapses to a line connecting the two point feet in case of double stance or to a point in the case of a single stance. As a result, the possible solution space becomes infeasible in the absence of contact moments.
This extension of the feasible region to degenerate cases is made numerically possible by assuming the presence of infinitesimal contact torques at the feet as constraints on the problem to render it feasible. In particular, we assume that the feet can exert a small torque component tangential to the contact surface plane $\tau_x$ and $\tau_y$, but no contact torque  orthogonal to the plane $\tau_z$. This corresponds to the case of feet with a small non-zero surface, able to adjust the location of the \gls{cop} within the contact surface. Such assumption is a numerical (heuristic) assumption introduced solely for the feasibility of the problem. We include these wrench components in the constraint (III.b) of Algorithm \ref{alg:IP}: we update the matrix $\mathbf{B}$ in \eqref{eq:force_constraint_matrix} to embed, for each contact $i$, not just the constraints on the contact forces (\ie linearized friction cone constraint $\mathbf{b}_i^{cone} \in \Rnum^{4 \times 3}$) but also a box constraint $\mathbf{b}_i^{box} \in \Rnum^{4 \times 2}$ on the contact torques $\tau_x, \tau_y$. The values $\tau_x^{lim}, \tau_y^{lim}$ represent the infinitesimal limits of the box constraint on the contact torque tangential to the surface plane in the foot location:

\begin{equation}\label{eq:IPmatrices}
\begin{aligned}
	&	 \mathbf{b}_i^{cone} = \mat{
		(\mathbf{t}_{1,i} - \mu_i \mathbf{n}_i)^T\\ (\mathbf{t}_{2,i} - \mu_i \mathbf{n}_i)^T \\ -(\mathbf{t}_{1,i} + \mu_i \mathbf{n}_i)^T \\ - (\mathbf{t}_{2,i} + \mu_i \mathbf{n}_i)^T},
	\quad \mathbf{b}_i^{box} = \mat{
		\tau_x^{lim} & 0\\ 0 & \tau_y^{lim} \\ -\tau_x^{lim} & 0 \\ 0 & - \tau_y^{lim}} \\	
	&\mathbf{B} = \diag\left(\mat{\mathbf{b}_1^{cone} & \mathbf{0}_{4\times2}\\ \mathbf{0}_{4\times3} & \mathbf{b}_1^{box}} \dots \mat{\mathbf{b}_{n_c}^{cone} & \mathbf{0}_{4\times2}\\ \mathbf{0}_{4\times3} & \mathbf{b}_{n_c}^{box}}\right)\in \Rnum^{8 n_c \times 5 n_c}
\end{aligned}
\end{equation}

\begin{figure}[tb]
\centering
\includegraphics[width=0.9\linewidth]{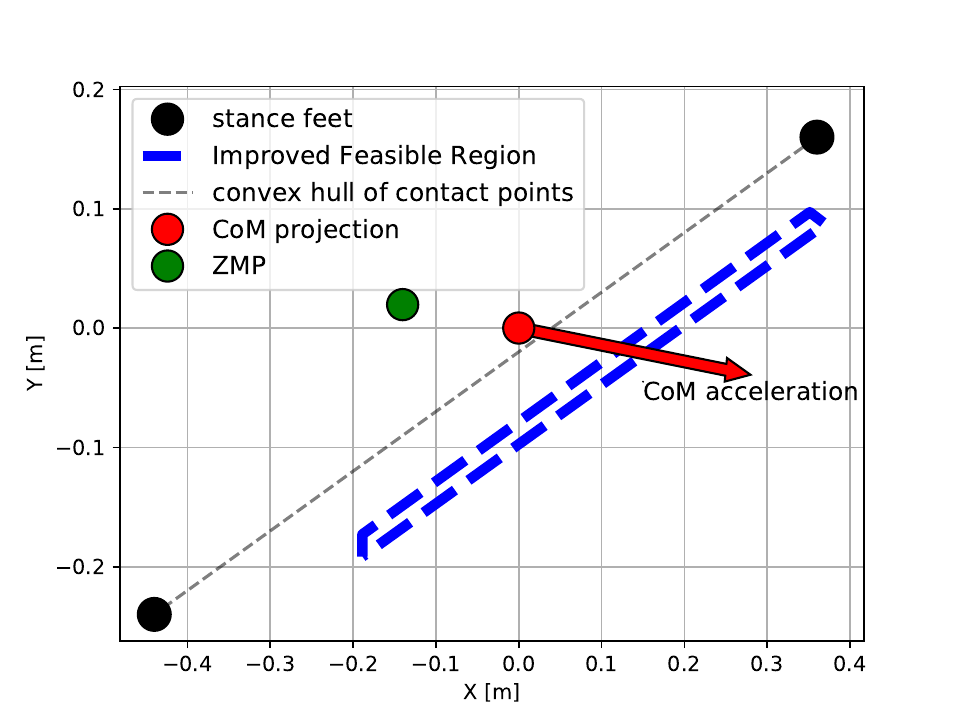}
\caption{\small The improved feasible region degenerates to a line during a trot, when only two feet are simultaneously in contact with the ground. 
	This segment is shifted forward in the same direction of the robot's acceleration. The finite width of the improved feasible region (blue) is due to the infinitesimal contact torques $\tau_x^{lim}$ and $\tau_y^{lim}$. \revisedtextthird{The robot is dynamically unstable in this scenario.}}
\label{fig:degenerate_regions}
\end{figure}

Because of the non-zero values of the contact torque limits $\tau_x^{lim}$ and $\tau_y^{lim}$,
the feasible region portrayed in Fig. \ref{fig:degenerate_regions} appears as a 
narrow stripe with finite area, although it should be regarded as a one-dimensional segment. 
Indeed, the feasible region in this double point-contact case corresponds to a segment 
whose length is determined by the robot's actuation limits. In presence of external 
wrenches acting on the platform, this segment will move away from the line 
connecting the two feet along the projection plane. 

In case of a single point contact, the feasible region will degenerate to a point 
which represents the only possible value of \gls{com} projection where the robot could balance 
the load acting on its trunk. Note that if the dynamic effects are considered, the feasible line will move back/forth when the robot accelerates backwards/forward, according to what is explained in Section \ref{sec:dynamics}.
This is exemplified in Fig. \ref{fig:degenerate_regions} which shows the feasible 
region during a trotting motion. The region is a straight segment and is shifted 
forward with respect to the supporting line, because the robot is accelerating forward. The \gls{zmp} (green point), 
instead, moves backwards in the opposite direction to the acceleration.
In future works, we plan to exploit this to perform fast turning maneuvers  to check the 
maximum feasible sideways inclination that can be achieved (\eg to compensate centrifugal forces).

\section{Reachable Region}
\label{sec:kin_lim}
So far the feasible region was defined as a region for which the 
frictional stability of the robot can be ensured 
without violating the joint-torque limits.
The inclusion of the effect of the joint-torque limits
%, through the set of feasible contact wrenches, 
has proved to be important in many cases. 
Once the torque-limits are considered, the limited leg workspace remains 
the next major restrictive factor for motion planning.
\revisedtext{This is particularly true in complex terrains, where the robot needs to have complex configurations that may result in joint-kinematic limits violations or leg singularities.} 
Kinematic limits are common, for instance, in linear actuators 
used in hydraulic quadrupeds, such as \gls{hyq}, where the piston stroke is limited.
%
%type of singularity
One type of singularity that could be of crucial 
importance to determine the workspace, is related 
to the loss of mobility due to the complete extension or 
retraction of one of the legs (\eg humanoid climbing stairs).
In fact, as it will be shown in this section, 
it often happens that, even if the \textit{feasible region} is  sufficiently large, 
yet the robot \gls{com} has a very limited reachable workspace.
Parallel robots in general, inherently  suffer from such an unfavorable workspace.

We, therefore, seek to extend the definition of the \textit{feasible region} to further 
incorporate the joint-kinematic limits and the manipulability of the robot. 
We first introduce the \textit{reachable region}, a two-dimensional level 
area representing the  \gls{com} reachable workspace. We present a simplified numerical 
approach that  computes a conservative approximation of the region. 
The method is designed to be \textit{efficient} and 
therefore  allows for \textit{online} motion planning and optimization.
Given a desired orientation, we determine the \textit{constant orientation workspace}: 
namely, the set of all possible \gls{com} locations that can be reached with a specified 
orientation without violating the joint-kinematic limits \cite{parallel}. 
To simplify the nomenclature, we refer to it as the \textit{reachable region}.
Given the kinematic nature of the problem, we can utilize the 
forward kinematic relations to map
the kinematic constraints of the robot (defined in the joint space) 
to the task-space (defined in the Cartesian space of the \gls{com}). 
Typically the forward kinematics for each branch in contact (\ie leg) is defined as:
\begin{equation}
{}^\mathcal{B}\mathbf{x}_{f_i} = f_i(\mathbf{q}_i), \quad \forall i= 1, ..., n_c
\label{eq:fk}
\end{equation}
mapping the joint angles $\mathbf{q}_i \in \mathbb{R}^{n_l}$ of branch $i$ to the position of 
the foot ${}^\mathcal{B}\mathbf{x}_{f_i} \in \mathbb{R}^{3}$ (expressed with respect to 
the body frame).
Assuming that the foot position with respect to the world frame ${}^\mathcal{W}\mathbf{x}_{f_i}$ is known, ${}^\mathcal{B}\mathbf{x}_{f_i}$ can be simply computed as
\begin{equation}
{}^\mathcal{B}\mathbf{x}_{f_i} = {}^\mathcal{B}\mathbf{R}_\mathcal{W} ({}^\mathcal{W}\mathbf{x}_{f_i} - \mathbf{c} ) + 
{}^\mathcal{B}\mathbf{c}
\label{eq:com_relation}
\end{equation}
where ${}^\mathcal{B}\mathbf{c}$ is the offset of the \gls{com} with respect to the body frame, and  $\mathbf{c}$ is the \gls{com} position with respect to the world frame.
Combining (\ref{eq:fk}) and (\ref{eq:com_relation}) and rewriting for $\mathbf{c}$, we obtain:
\begin{equation}
\mathbf{c} = \mathbf{F}_i(\mathbf{q}_i, {}^\mathcal{W}\mathbf{x}_{f_i}, {}^\mathcal{B}\mathbf{R}_\mathcal{W}), \quad  \forall i= 1, ..., n
\label{eq:F_relation}
\end{equation}
where $\mathbf{F}_i$ is defined as:
\begin{equation}
\mathbf{F}_i(\mathbf{q}_i, {}^\mathcal{W}\mathbf{x}_{f_i}, {}^\mathcal{B}\mathbf{R}_\mathcal{W}) = 
{}^\mathcal{W}\mathbf{x}_{f_i} - {}^\mathcal{W}\mathbf{R}_\mathcal{B} ( f_i(\mathbf{q}_i) - {}^\mathcal{B} \mathbf{c} )
\label{eq:F}
\end{equation}
Therefore, for a given foot position ${}^\mathcal{W}\mathbf{x}_{f_i}$ and  trunk orientation ${}^\mathcal{W}\mathbf{R}_B$, (\ref{eq:F_relation}) provides a relationship between the  joint-space angles of each leg and the \gls{com}  task-space position. We assume the feet do not move during contact. This is enforced by the \gls{wbc} used in our framework (see Fig. \ref{fig:crawl_diagram}) \cite{wbc}. Therefore, for a \gls{com}  position ${}^\mathcal{W}\mathbf{x}_{com}$ to be reachable, there must exist joint angles $\mathbf{q}_i$, satisfying \eqref{eq:F_relation}, for each leg $i$ such that:
\begin{enumerate}
\item $\underline{\mathbf{q}}_i \leq \mathbf{q}_i \leq \bar{\mathbf{q}}_i$ \label{lim}
\item $J_i(\mathbf{q}_i) = [\partial f_i(\mathbf{q}_i)/\partial \mathbf{q}_i] \quad \text{is full rank}$ \label{jacob}
\end{enumerate}
where $\underline{\mathbf{q}}_i$ and $\bar{\mathbf{q}}_i$ are the minimum and maximum joint angle limits, respectively and $\leq$ is an element-wise relational operator. 

We can therefore utilize (\ref{eq:F_relation}) (we drop the explicit dependence 
on ${}^{\mathcal{W}}\mathbf{x}_{f_i}$ and ${}^\mathcal{W}\mathbf{R}_B$ that are input parameters, to lighten the notation), 
along with conditions (\ref{lim}) and (\ref{jacob}) defined above, to define the \textit{reachable region} as:
\begin{equation}
\begin{aligned}
	&\mathcal{Y}_{r} = \Big\{\mathbf{c}_{x y} \in \mathbb{R}^{2} | \quad  \exists \mathbf{q}_i \in \mathbb{R}^{n_l} \text { s.t. } \quad (\mathbf{c}_{x y}, \mathbf{q}_i) \in \mathcal{Q}\Big\}
\end{aligned}
\label{eq:reach_def}
\end{equation}
where:
\begin{equation}
\begin{aligned}
	&\mathcal{Q} = \Big\{\mathbf{q}_i \in \mathbb{R}^{n_l}, \mathbf{c}_{x y} \in \mathbb{R}^{2} | \text { s.t. } \quad \mathbf{c}_{x y} = \mathbf{P}_{xy} \mathbf{F}_i(\boldsymbol{q}_i),\\
	& \underline{\boldsymbol{q}}_i \leq \boldsymbol{q}_i \leq \bar{\boldsymbol{q}}_i, 
	\quad \text{row-rank}(J_i(\mathbf{q}_i)) = n_l \quad \forall i = 1,..., n_c \Big\}
\end{aligned}
\label{eq:joints_set}
\end{equation}
where only the legs in contact are considered. It is important to note that such set can be composed from 
the intersection of pairs of concentric circles \cite{stewart_analytic}. This in general results in a 
non-convex set. The problem of finding such set accurately is difficult and time consuming. Various 
techniques have been proposed to determine the workspace of manipulators by using analytic, geometric, 
or numerical approaches. Most analytic and geometric methods can turn the 
analysis of the geometry very complex or can be specific to only one platform. 
We therefore employ a numerical approach that provides an approximation of the region 
smartly designing it to remain efficient for any generic platform.
Numerical methods mostly either sample the joint-space and utilize the forward kinematics or, 
conversely, sample the task-space and utilize the inverse kinematics. In the case of quadrupeds, the 
dimension of the joint-space can be  large (12-dimensional in the case of most robots). Therefore we choose to 
utilize the inverse kinematics to determine the reachable region.

Algorithm \ref{alg:reach_alg} describes the procedure developed to compute the region. A similar 
algorithm was developed in \cite{stewart_numerical}, and was used to evaluate the workspace of a 
Stewart platform based machine tool. We further apply a modification to increase the robustness and the 
performance.
Inspired by ray-casting algorithms, a discretized search is 
done iteratively in ordered directions 
along polar coordinates $(\rho, \theta)$ starting from the current  \gls{com} projection.
This generates a 2D polygon whose vertices are ordered and belong to the boundary of the reachable region, therefore
representing a \textit{polygonal} approximation of the said region.
For the sake of simplicity, for the remainder of this paper, 
we will refer to the reachable region $\mathcal{Y}_{r}$ as its polygonal approximation.

\begin{figure}[tb]
\centering
\begin{subfigure}{7cm}
	\includegraphics[width=1\textwidth]{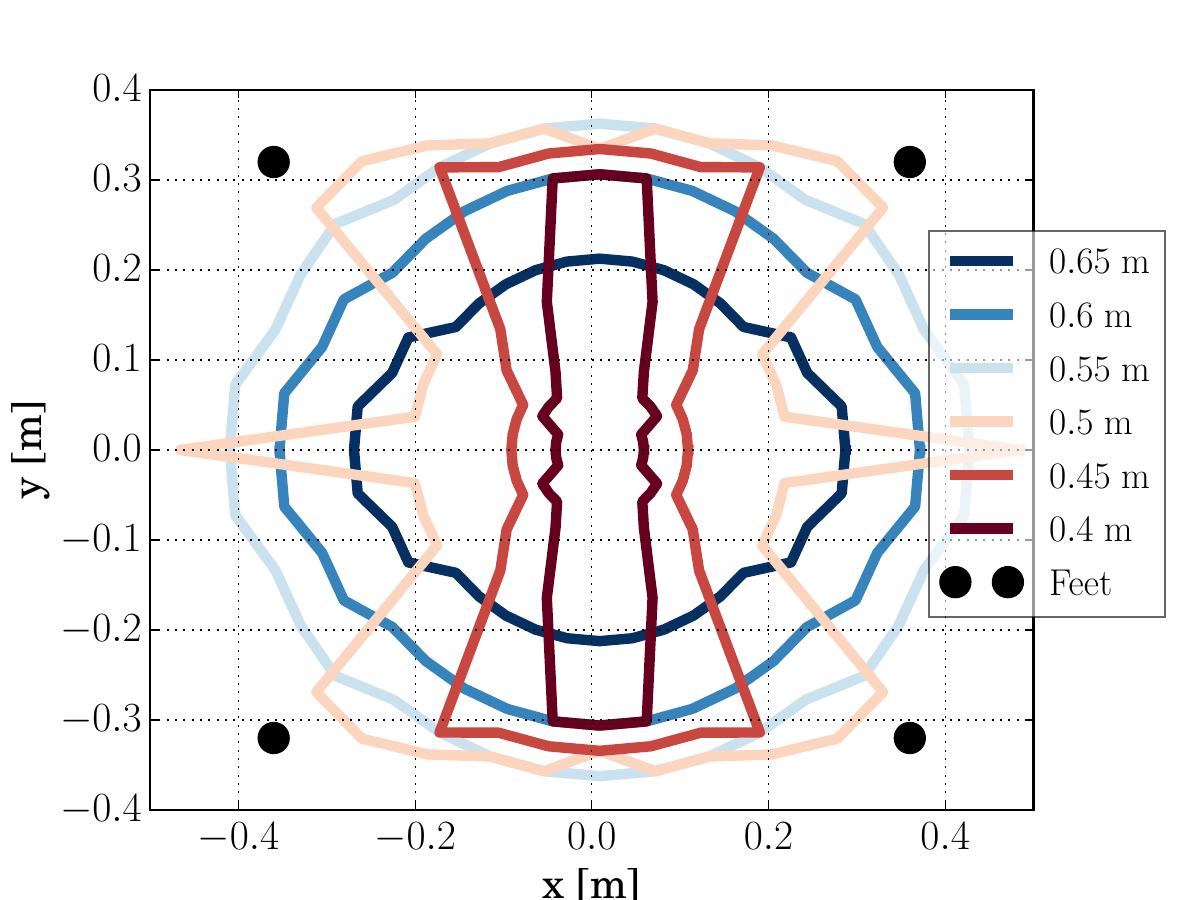}
	\caption{Four stance feet.}
\end{subfigure}\\
\begin{subfigure}{7cm}
	\includegraphics[width=1\textwidth]{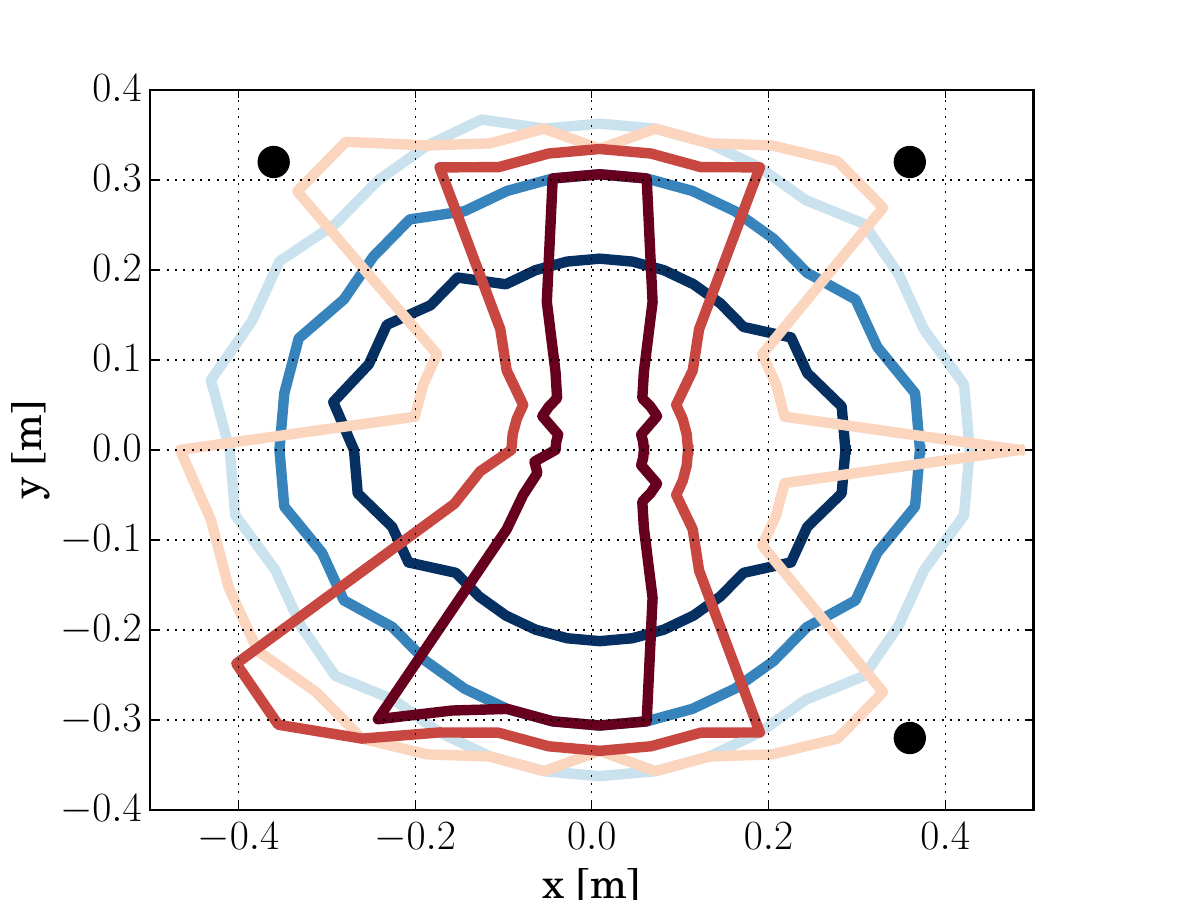}
	\caption{Three stance feet.}
\end{subfigure}
\caption{\small Different evaluations of the reachable region at different \gls{hyq} \gls{com} heights.}
\label{fig:reach_region_height}
\end{figure}

\begin{figure}[tb]
\centering
\includegraphics[width=0.38\textwidth]{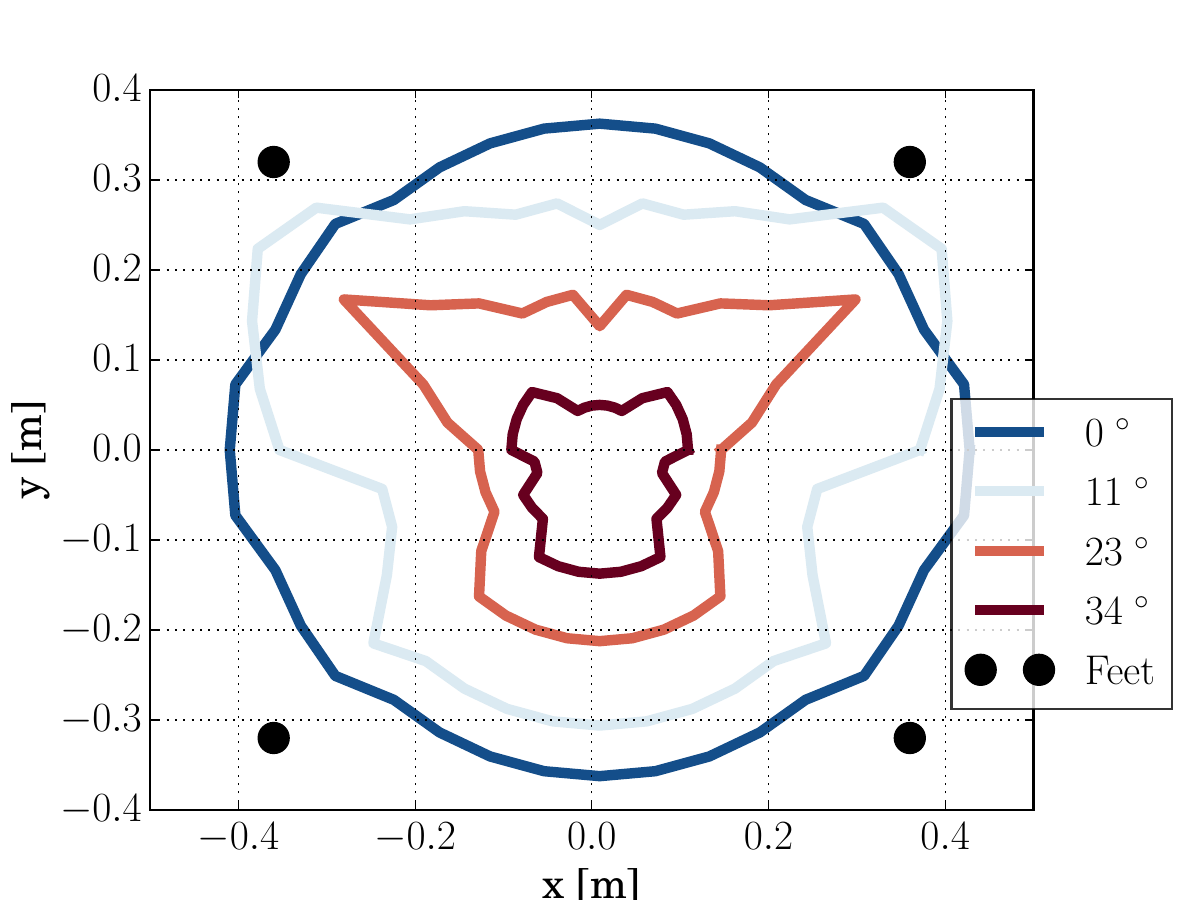}
\caption{\small Different evaluations of the reachable region at different \gls{hyq} roll orientations.}
\label{fig:reach_region_orient}
\end{figure}

Each ray along some direction $\mathbf{a}_i$ finds the 
farthest point $\bm{\nu}^*_{xy}$ that \revisedtext{still} belongs to the region. 
By construction, this point belongs to the boundary of the region 
and the problem of computing it can be stated, 
utilizing the inverse kinematics, as:
\begin{gather}
\underset{\bm{\nu}_{xy}} {\text{max~}} \mathbf{a}^T_i\bm{\nu}_{xy}\\
\label{eq:LP}
\text{s.t. $\forall i = 1,...,n_c$:}   \hspace{65pt}  \nonumber  \\
\hspace{25pt} \mathbf{q}_i = \mathbf{\bar{F}}_i(\bm{\nu}_{xy})
\label{eq:ik}\\
\hspace{25pt} \underline{\mathbf{q}}_i <\mathbf{q}_i <\bar{\mathbf{q}}_i \label{eq:lim} \\
\hspace{30pt} \sigma_{\min }\left\{J\left(\mathbf{q}^{k}\right)\right\} > \sigma_{0} 
\label{eq:singl}
\end{gather}
The relation (\ref{eq:ik}) represents the kinematic constraint 
in (\ref{eq:joints_set}) reformulated in terms of the inverse kinematics. $\mathbf{\bar{F}}_i$, therefore, is defined as:
\begin{equation}
\mathbf{\bar{F}}_i(\bm{\nu}_{xy}) = f^{-1}_i[{}^\mathcal{B}\mathbf{R}_\mathcal{W} ({}^\mathcal{W}\mathbf{x}_{f_i} - \mathbf{P}^T_{xy} \bm{\nu}_{xy} - \mathbf{P}^T_{z}c_z) +  {}^\mathcal{B}\mathbf{c}] 
\label{eq:inverse_F}
\end{equation}
where $f^{-1}_i$ refers to the inverse kinematics mapping.
It is important to note from (\ref{eq:inverse_F}) that for 
specific feet positions, the location of each $\bm{\nu}^*_{xy}$
(and accordingly the resultant region) is influenced by the height $c_z$ and the orientation ${}^\mathcal{W}\mathbf{R}_B$ of the robot.
A simple  check for the presence of a singularity is done in 
(\ref{eq:singl}), where $\sigma_{\min }$ is the smallest singular value and $\sigma_{0}$ 
is a small value of choice.
Due to the non-linearity of constraints \eqref{eq:ik} and \eqref{eq:singl} the problem 
cannot be casted as a linear program (LP) and we employ a ray-casting approach for the solution.
A bisection search could be utilized to speed up the search for $\bm{\nu}^*_{xy}$.
We first perform an evenly distributed search along the selected direction $\mathbf{a}_i$, with steps $\Delta \rho$, to find both the last point inside the 
region and the first point outside.
These correspondingly generate the interval $[\rho - \Delta \rho, \rho]$ where $\bm{\nu}^*_{xy}$ lies in. A fast bisection search is then executed on this interval to find $\bm{\nu}^*_{xy}$ while making sure it is within an error of [0, $-\Delta \rho_{\text{min}}$] from the boundary of the actual workspace. The function \textit{isReachable($\rho$)}, used in Algorithm \ref{alg:reach_alg}, computes the inverse kinematics of a CoM position and checks if that position is reachable:
\begin{algorithmic}
\State \textbf{isReachable($\rho$)}:
\State \hspace{40pt} $\bm{\nu}_{xy} \gets \mathbf{c}_{xy} +  \rho \mathbf{a}$
\State \hspace{45pt}$\mathbf{q_i} = \mathbf{\bar{F}}_i(\bm{\nu}_{xy})$
\State \hspace{40pt} \Return true if $\mathbf{q_i}$ satisfies (\ref{eq:lim}) \& (\ref{eq:singl})
\end{algorithmic}
\begin{algorithm}
%\caption{Reachability region computation}
\caption{Iterative discretized ray-casting algorithm}
\label{alg:reach_alg}
\begin{algorithmic}[1]
	\State \textbf{Input: } $\mathbf{c}_{xy}, c_z, {}^W\mathbf{R}_B, \mathbf{p}_1, ..., \mathbf{p}_{n_c},
	\underline{\mathbf{q}}_1, \bar{\mathbf{q}}_1,...,   \underline{\mathbf{q}}_{n_c}, \bar{\mathbf{q}}_{n_c}  $
	\State \textbf{Result: } reachable region $\mathcal{Y}_{r}$
	\State \textbf{Initialization}: $\bm{\nu}_{xy} = \mathbf{c}_{xy}$, $\mathcal{Y}_r \gets \{ \}$
	\For{ $\theta = 0$ to $2\pi$}
	\State Compute direction: $\mathbf{a}_i=\mat{\cos\theta & \sin \theta & 0}^T$
	\Statex
	%		\Statex \hspace{12pt} \textbf{Find vertex}:
	\Statex \hspace{12pt} \textit{Find the first bisection interval}:
	\While{\textit{isReachable($\rho$)}}
	\State $\rho \gets \rho + \Delta \rho$
	\EndWhile
	\Statex
	\Statex \hspace{12pt} \textit{Bisection search}:
	\State $\Delta \rho \gets \frac{\Delta \rho}{2}$
	% Between \rho - \Delta \rho and \rho
	\While{$\Delta \rho \geq \Delta \rho_{\text{min}}/2$}
	\If{\textit{isReachable($\rho$)}}
	\State $\rho \gets \rho + \Delta \rho$
	\Else
	\State $\rho \gets \rho - \Delta \rho$
	\EndIf
	%		\Until{$|\rho| < \rho_{\text{min}}$}
	\State $\Delta \rho \gets \frac{\Delta \rho}{2}$
	\EndWhile
	\Statex
	\If{last $\bm{\nu}_{xy}$ not \textit{isReachable($\rho$)}}
	\State $\rho \gets \rho - \Delta \rho_{\text{min}}$
	\State $\bm{\nu}_{xy} \gets \mathbf{c}_{xy} +  \rho \mathbf{a}$
	\EndIf
	\State $\mathcal{Y}_{r} \cup \left\{\bm{\nu}^*_{xy}\right\}$
	\EndFor
	\State \Return $\mathcal{Y}_{r}$
\end{algorithmic}
\end{algorithm}
Each vertex $\bm{\nu}^*_{xy}$ is added to the vertex description  
$\mathcal{Y}_{r}$ such that the (non-convex) hull of the ordered set of
vertex becomes an approximation of the real reachable region 
(see Fig. \ref{fig:reach_region_height} and \ref{fig:reach_region_orient}).
The algorithm stops when a step smaller than $\Delta \rho_{min}/2$ set by the user, is reached\footnote{$\Delta \theta = 10\degree$ and $\Delta \rho_{\text{min}} = 0.03 m$ are used for the shown figures. For a faster computation during planning, $\Delta \theta = 20\degree$ was sufficient for the simulation and experimental results.}\footnote{The large peaks in the edges of the reachable regions in Fig. {\ref{fig:reach_region_height} and \ref{fig:reach_region_orient}} are due to the physical nature of the workspace. Smaller rough edges along the boundary of the regions are due to the discretization used.}.

A key assumption taken in the algorithm is that the center of the \textit{reachable region} 
is the \textit{current} \gls{com} location. This speeds up a 
necessary first step of searching for an approximate center to start the algorithm from.
Moreover, this provides better boundary precision when determining the 
boundary of the region that is closer to the \gls{com} position, presenting a safer analysis.
As a consequence, the dependence of the algorithm from $\mathbf{c}_{xy}$, 
only influences the \textit{accuracy} of the generated region.
A disadvantage of such choice is the 
inability to compute the region if the robot is already in an out-of-reach configuration.
Nevertheless, given that the locomotion planning shall be done in coherence with the reachable
region (see Section \ref{sec:com_planning}), the trajectory of the \gls{com} shall always remain inside the region. The Appendix provides a further discussion of the nature of the \gls{com} workspace and the reachable region.

On the other hand, it is important also to consider the effect of the robot  
height $c_z$ and orientations ${}^\mathcal{W}\mathbf{R}_\mathcal{B}$ on the reachable region.
In fact,  different evaluations of the reachable region, 
presented in Fig. \ref{fig:reach_region_height} and \ref{fig:reach_region_orient},
show that the size, positioning, shape, and convexity of the reachable region can differ greatly
at different $c_z$ and ${}^\mathcal{W}\mathbf{R}_\mathcal{B}$.
Unsurprisingly, one can observe that the region tends to become smaller at high and low 
heights, since the legs have in general less mobility when fully extended or retracted.
Furthermore, a deviation from the default horizontal orientation 
results in smaller regions and could additionally skew the shape of the region towards one side.
In both cases, at certain  configurations, the convexity of the region can be significantly affected.
Such insight is greatly useful in situations where planning 
needs to be performed in \revisedtextthird{rough terrains}.

\section{The Improved Feasible  Region}
\label{sec:improved}
The reachable region can be seen as a projection of the high-dimensional
convex set $\mathcal{Q}$ onto a 2D subspace.
Henceforth, with the feasible region and the reachable region defined on the same plane, 
one could extend the definition of the feasible region to further 
include the \gls{com} positions that are also reachable.
In other words, this would present a comprehensive 2D region of all 
the feasible \gls{com} positions $\mathbf{c}_{xy}$ that satisfy the friction constraints, 
the joint-torque constraints, and the joint-kinematic constraints simultaneously.
We can therefore define an \revisedtext{\textit{improved}} \textit{feasible region} as:
\begin{equation}
\begin{aligned}
	\mathcal{Y}_{far} = \Big\{\mathbf{c}_{x y} \in \mathbb{R}^{2} |& \quad  \exists \mathbf{f}_i \in \mathbb{R}^{m n_c}, \mathbf{q}_i \in \mathbb{R}^{n_l} \text { s.t. } \\ &(\mathbf{c}_{x y}, \mathbf{f}_i) \in \mathcal{C} \cap \mathcal{A}, \quad (\mathbf{c}_{x y}, \mathbf{q}_i) \in \mathcal{Q}\Big\}
\end{aligned}
\end{equation}
Given that $\mathcal{C} \cap \mathcal{A}$ and $\mathcal{Q}$ are defined on different spaces, $\mathcal{Y}_{far}$ can therefore be obtained by computing
the feasible region $\mathcal{Y}_{fa}$ (projecting $\mathcal{C} \cap \mathcal{A}$) and the reachable region $\mathcal{Y}_{r}$ (projecting $\mathcal{Q}$) separately,  then considering the intersection of the two regions (see Appendix for additional details).
Therefore we can define the \revisedtext{\textit{improved}} \textit{feasible region} as:
\begin{equation}
\mathcal{Y}_{far} = \mathcal{Y}_{fa} \cap \mathcal{Y}_{r}
\end{equation}
Finally, differently from the $\mathcal{Y}_{fa} $ region, that took 
into account only friction and joint-torque constraints, 
the \revisedtext{improved} feasible region $	\mathcal{Y}_{far}$ will be non-convex because the reachable 
region is non-convex (given that the set produced from the intersection between a 
convex set and a non-convex set is non-convex). 
In Table \ref{tab:regions} we summarize the type of regions 
introduced together with the correspondent constraints.
\begin{table}[h!]
\begin{center}
	\resizebox{0.5\textwidth}{!}{
		\begin{tabular}{| c | c | c |}
			\hline	\hline
			\textbf{Name} &  \textbf{Symbol}  & \textbf{Constraints} \\ 
			\hline
			Friction R. (\cite{Bretl2008a}) & $\mathcal{Y}_{f}$  & Friction  \\
			\hline
			Feasible R. (\cite{orsolino19tro}) &  $\mathcal{Y}_{fa}$ &  Friction / Joint-torque \\
			\hline
			Reachable R. (this paper) & $\mathcal{Y}_{r}$  & Kinematic \\ 
			\hline				
			Improved Feasible R. (this paper) &  $\mathcal{Y}_{far}$  &  Friction / Joint-torque / Kinematic \\
			\hline	\hline
		\end{tabular}	
	}
	\caption{Types of regions}
	\label{tab:regions}
\end{center}
\end{table}

\section{Trajectory Planning} 
\label{sec:planning}
\begin{figure}[tb]
	\centering
	\includegraphics[width=0.4\textwidth]{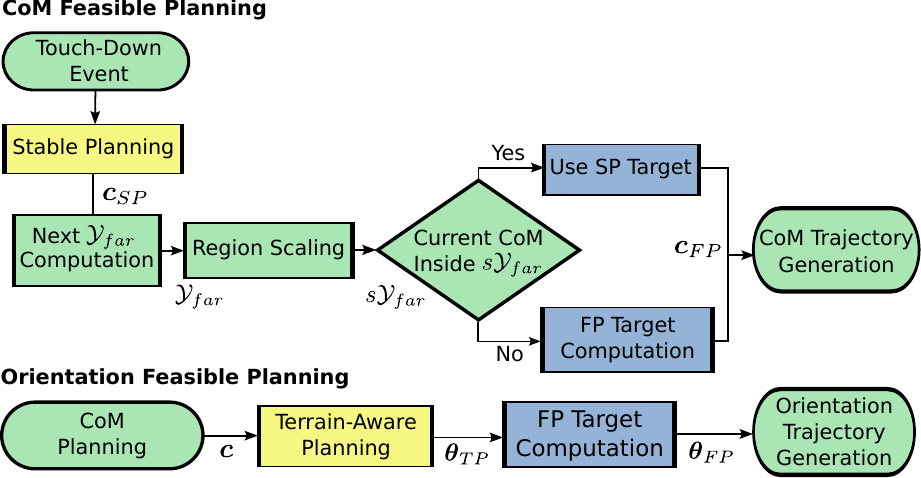}
	\caption{\small \revisedtextthird{Flowcharts illustrating the trajectory planning algorithms. Top: CoM FP algorithm. Bottom: Orientation FP algorithm.}}
	\label{fig:planning_flowchart}
\end{figure}
\subsection{CoM planning strategy} 
\label{sec:com_planning}
In this subsection we improve the \gls{com} planning strategy developed for \textit{crawl} 
gaits described in our previous work \cite{Focchi2020}, by exploiting the proposed definition of \revisedtext{the} improved feasible region (see Fig. \ref{fig:crawl_diagram} \revisedtextthird{and \ref{fig:planning_flowchart}}). \revisedtextthird{We will denote the method used in \cite{Focchi2020} as the \textit{stable planning (SP)} strategy and our improved one as the \textit{feasible planning (FP)} strategy}.
To simplify the planning framework, we assume a quasi-static motion: during a crawl cycle, 
the robot motion is divided into \textit{swing phases}, where only one foot is 
allowed to swing while the robot trunk is kept stationary,
and \textit{move-body phases}, where all feet are in stance 
and the trunk is moved to a target location and orientation. 
A pre-defined foot sequence is used\footnote{The default locomotion
sequence for crawl is: \gls{rh}, \gls{rf}, \gls{lh}, \gls{lf}}. Note that it is possible to extend the strategy to a more dynamic gait by designing \gls{com} trajectories that are consistent with the extension shown in Section \ref{sec:dynamics}. Therefore, one would need to ensure the trajectory is consistent with the dynamic feasible region, and that the accelerations are consistent with the desired ones that the region was computed for.
Through the use of the \revisedtext{improved} feasible region we will improve the 
\revisedtextthird{SP} behavior adding guarantees on the 
physical feasibility.
The  feasible region is utilized to plan a \gls{com} trajectory for 
the move-body phase such that in the following swing phase, 
\ie when only three feet are in stance, the \gls{com} target remains feasible.

This phase (also labeled as three-\revisedtext{contact} phase)
is the most critical in terms of stability (the friction region is typically smaller)
and actuation capability, as only three legs support the whole robot weight and the other possible external wrenches.
After each touch-down (\ie at the start of a move-body phase),
the next  feasible region $\mathcal{Y}_{far}$  is computed, 
based on the future three stance legs (known from the foot 
sequence). A \revisedtextthird{FP} target \gls{com} 
position, using the criterion explained below, is then chosen.
In such manner, the feasibility is ensured when 
the next swing foot is lifted and the robot is only supported by three feet.
A quintic polynomial trajectory for the \gls{com} is generated 
linking the current \gls{com} position with the chosen 
target and is tracked during the move-body phase in progress. \revisedtextthird{Figure \ref{fig:planning_flowchart} provides a flowchart of the planning algorithms.}
As mentioned in Section \ref{sec:recap_feasible}, 
the Jacobians used to evaluate the force polytopes of the 
contact legs make the  feasible region  configuration-dependent.
Therefore, the region should be recomputed for each CoM location along the planned trajectory.
To simplify the planning problem, we instead evaluate the region using the leg Jacobians computed 
at a configuration corresponding to the \revisedtextthird{SP} CoM target. The use of a single Jacobian can be further justified by the analysis done in \cite{orsolino18ral}, which showed
that the variation of the Jacobian, around a specific configuration,
has negligible effect on the contact forces.
To introduce a level of robustness against uncertainties, 
the planning of the target is  done considering
a \textit{scaled} version of the  feasible 
region $s\mathcal{Y}_{far}$ with a tunable scaling coefficient $s \in (0, 1)$.

The procedure is devised as follows: if the current \gls{com} projection 
$\textbf{c}_{xy}$ (onto the region plane)\footnote{In the accompanying video, 
the projected regions are illustrated at the feet level just for visualization purposes. 
However, the computation of the regions has been performed at the level of the \gls{com}.}
is inside $s\mathcal{Y}_{far}$, feasibility is already guaranteed and the 
target  \gls{com}  position is chosen to be the current one to minimize unneeded motion.
Otherwise, we proceed to select the point on the boundary of $s\mathcal{Y}_{far}$, 
that is closest to the target computed using the \revisedtextthird{SP}. This allows the motion to:
(1) be as close as possible to the \revisedtextthird{SP} target; thus benefiting from its proven reliable practical effectiveness \cite{Focchi2020};
(2) formally fulfill the feasibility requirements; and
(3) achieve a desired level of robustness (tunable by the shrinkage factor $s$). Remaining close to the \revisedtextthird{SP} target, also allows to
(4) maintain the \textit{local} validity of the \textit{feasible region} 
(the Jacobian was evaluated for the \revisedtextthird{SP} target position).

Furthermore, the scaling of a convex polygon can be performed 
through an affine transformation with respect to the Chebyshev center or the centroid (see \cite{orsolino19tro}).
For non-convex polygons, this problem is harder. 
One solution is to use inward polygon offsetting. However, this is not yet fast enough for online planning and we have noticed that, for this purpose, scaling the feasible region with respect to its centroid provides satisfactory results. For a more detailed discussion, refer to the Appendix.

\subsection{Optimization of trunk orientation to maximize joint range}
\label{sec:orient_planning}
%
%%%%
Upon planning a \gls{com} trajectory, our previous planning approach \cite{Focchi2020} \revisedtextthird{also}
computes a target trunk orientation (roll and pitch) to be attained during the move-body phase.
This target is chosen to be aligned with the inclination of the \textit{terrain plane} 
which is  estimated in \cite{Focchi2020} via fitting an averaging plane through the stance feet. \revisedtextthird{We will denote the method of \cite{Focchi2020} as the \textit{terrain-based planning (TP)} strategy and our improved one as the \textit{feasible planning (FP)} strategy}.

This \revisedtextthird{TP strategy} aims at bringing the legs as close as possible to the 
middle of their workspace in order to avoid the violation of the kinematic limits.
For instance, if the robot walks up a ramp, keeping a horizontal posture will lead the back legs 
to extend and the front ones to retract, risking a kinematic limit violation in some of the joints. 
However, for rough terrains, where the feet are located on distant non coplanar surfaces, this might not be sufficient.
In such cases, it can happen that some legs become more extended/retracted than others, \revisedtext{as will be illustrated in Section \ref{sec:rough_terrain_exp}}.

Examining the effect of the trunk orientation on the region in Section \ref{sec:kin_lim},
we can exploit the region to guide the choice of the orientation that best encloses 
the whole \gls{com} trajectory chosen in Section \ref{sec:com_planning}.
In particular, we choose to optimize the orientation to maximize the \textit{minimum} 
distance between the trajectory and the boundary of the region during the move-body phase.
This not only attempts to ensure the inclusion of the \textit{whole} trajectory in the region, 
but also tries to keep it away from the boundary as much as possible, thus increasing robustness.
In case multiple orientations result in similar distances, 
we opt for the one that maximizes the area of the region.
Optimizing for the orientation allows the robot to be less conservative 
in its movements and to achieve more complex configurations on rough terrains. 
In other words, we make sure that \textit{each leg has a minimum 
distance from the limits of its workspace}, as opposed to the previous \revisedtextthird{TP} 
approach that \revisedtextthird{handles all the legs collectively to better match the terrain inclination estimate}.

To reduce the size of the problem, it is necessary to initialize 
the search space around some solution.
As mentioned above, the \revisedtextthird{TP} orientation provides an elementary, yet satisfactory, behavior in many cases.
Accordingly, we choose to sample the orientation space around the \revisedtextthird{TP orientation}. 
Furthermore, we only optimize for the pitch and roll angles, since the 
yaw angle is computed to keep the base aligned with the locomotion direction. 

Note that this orientation planning strategy aims to improve upon the \gls{com} 
planning strategy described in Section \ref{sec:com_planning} 
and does not necessarily guarantee feasibility on its own;
a \gls{com}  target that is highly unfeasible for the 
default orientation is very likely to remain unfeasible for any other possible \textit{better} orientation.
For this reason, we choose to perform the \gls{com}  planning strategy in Section \ref{sec:com_planning} 
(computed at the default orientation) \textit{before}  optimizing for the orientation.

\section{Assumptions Summary}
\label{sec:assumptions}
As mentioned in the last sections, several assumptions were made during the computation of the algorithms presented in this paper. Table \ref{tab:assumptions} provides a summary of these assumptions, their types and purposes, and where it was discussed in the paper.
\begin{table}[h!]
\begin{center}
	\resizebox{0.5\textwidth}{!}{
		\begin{tabular}{| c | c | c | c | c|}
			\hline	\hline
			& \textbf{Assumption} & \textbf{Type} & \textbf{Purpose} & \textbf{Sections}\\ 
			\hline
			1 & Friction  pyramids & M & \revisedtextthird{CE} & \ref{sec:recap_feasible} \\
			\hline
			2 & Feet in contact are stationary & M & \revisedtextthird{PS} & \ref{sec:kin_lim} \\
			\hline
			3 & \revisedtextthird{Tangential c}ontact moments $\neq$ zero & %Deal with support polygon degeneration
			\revisedtextthird{M} & \revisedtextthird{SF} & \ref{sec:degenerate} \\ 
			\hline				
			4 & Approximate center of reachable region & A & \revisedtextthird{CE} & \ref{sec:kin_lim}, Appx.A \\
			\hline				
			5 & Scaling of region based on centroid & A & \revisedtextthird{CE} & \ref{sec:com_planning}, Appx.C \\
			\hline				
			6 & Single leg Jacobians for planning & M & \revisedtextthird{PS} & \ref{sec:com_planning} \\
			\hline
			7 & Quasi-static motion planning & \revisedtextthird{M} & \revisedtextthird{PS} & \ref{sec:com_planning} \\
			\hline	\hline
		\end{tabular}
	}
	\caption{List of assumptions used in the calculation of the improved feasible region and during planning, along with their types and purposes. The abbreviation of the types are as follows: M - Modelling and A - Algorithmic. \revisedtextthird{The abbreviation of the purposes are as follows: CE - Computation Efficiency, PS - Planning Simplicity, and SF - Solution Feasibility.}}
	\label{tab:assumptions}
\end{center}
\end{table}

\revisedtextthird{We classify the assumptions made to be of \textit{modelling} or \textit{algorithmic} type. \textit{Modelling} assumptions relate to the model of the robot or the model of the interaction of the robot with the environment. \textit{Algorithmic} assumptions relate to the definition of the algorithms that compute the regions or the planning target.}

\revisedtextthird{Each assumption was made for the purpose of either having \textit{computational efficiency}, \textit{planning simplicity}, or \textit{solution feasibility}.
\textit{Computation efficiency} refers to assumptions that are made for the purpose of avoiding large computation times. Other assumptions were made to demonstrate the capabilities of the improved feasible region while attempting to avoid complicating the planning method, i.e. \textit{planning simplicity}. \textit{Solution feasibility} refers to assumptions that are made for the purpose of adapting a solution method to our problem.}

\revisedtextthird{Some assumptions can be chosen to be relaxed as required. Assumptions that are made for the purpose of computation efficiency (i.e., 1, 4, and 5) can be dropped at the cost of higher computation times (possibly becoming unsuitable for online planning). Assumptions 6 and 7 can be relaxed and subsequently a more involved planning strategy is needed. Assumption 2 is assumed to be enforced by the control architecture and Assumption 3 is necessary with our solution method.}

\section{Simulation Results}
\label{sec:simulations}
To demonstrate the capability of the proposed improved feasible region, 
we devised some challenging scenarios that the robot 
has to traverse, designed to best illustrate the region's features.
Under such scenarios, we show the superior performance of planning based on
the  improved feasible region compared to heuristics \revisedtextthird{of \cite{Focchi2020}}.

All the presented simulations and experiments are shown in the accompanying video.
The generation of the projected regions is done in 
Python 2.7\footnote{Source code available at \hyperref[github.com/abdelrahman-h-abdalla/jet-leg]{github.com/abdelrahman-h-abdalla/jet-leg}.}. \revisedtext{Table \ref{tab:time} shows a summary of the computation times of the different stages of the planning}\footnote{We expect a decrease in the computation time upon 
performing the computation in C++, \eg using  Cython \cite{cython}.}.
\revisedtext{The computer is equipped with an i7-8700 3.2 GHz processor and 16GB of RAM.}
Whenever a multitude of  regions needs to be computed 
(as in the case of the optimization of the trunk orientation)
we make use of the parallelism capabilities of our CPU 
using the \textit{multi-processing} module in Python.
The regions are sent  via a ROS node to our locomotion planner,
that runs in a ROS environment. The \gls{wbc} runs at 250 $Hz$.

\begin{table}[h!]
\begin{center}
	\begin{tabular}{|l | c | c | c |}
		\hline	\hline
		\textbf{Stage} &  \multicolumn{3}{c|}{\textbf{Computation Time}} \\ 
		\cline{2-4 }
		& 2-contacts & 3-contacts & 4-contacts \\
		\hline
		Feasible region & 5 $ms$ & 9 $ms$ & 14 $ms$ \\
		\hline
		Reachable region & 25 $ms$ & 28 $ms$ & 29 $ms$ \\
		\hline
		Intersection $\mathcal{Y}_{far}$ & \multicolumn{3}{c|}{0.3 $ms$} \\ 
		\hline				
		Region scaling & \multicolumn{3}{c|}{0.005 $ms$} \\
		\hline						
		Target planning & \multicolumn{3}{c|}{ 0.03 $ms$} \\
		\hline						
		Total feasible planning & \multicolumn{3}{c|}{18 $Hz$ (worst case)} \\
		\hline						
		Whole-body controller & \multicolumn{3}{c|}{ 250 $Hz$ (worst case)} \\
		\hline	\hline
	\end{tabular}
	\caption{\small Average computation time for each stage of planning\revisedtextthird{, using the SP target as an initialization for the required stages}.}
	\label{tab:time}
\end{center}
\end{table}
\subsection{Walk in cluttered environment}
\label{sec:exp_cluttered}
\begin{figure}[tb]
\centering
\includegraphics[width=0.4\textwidth]{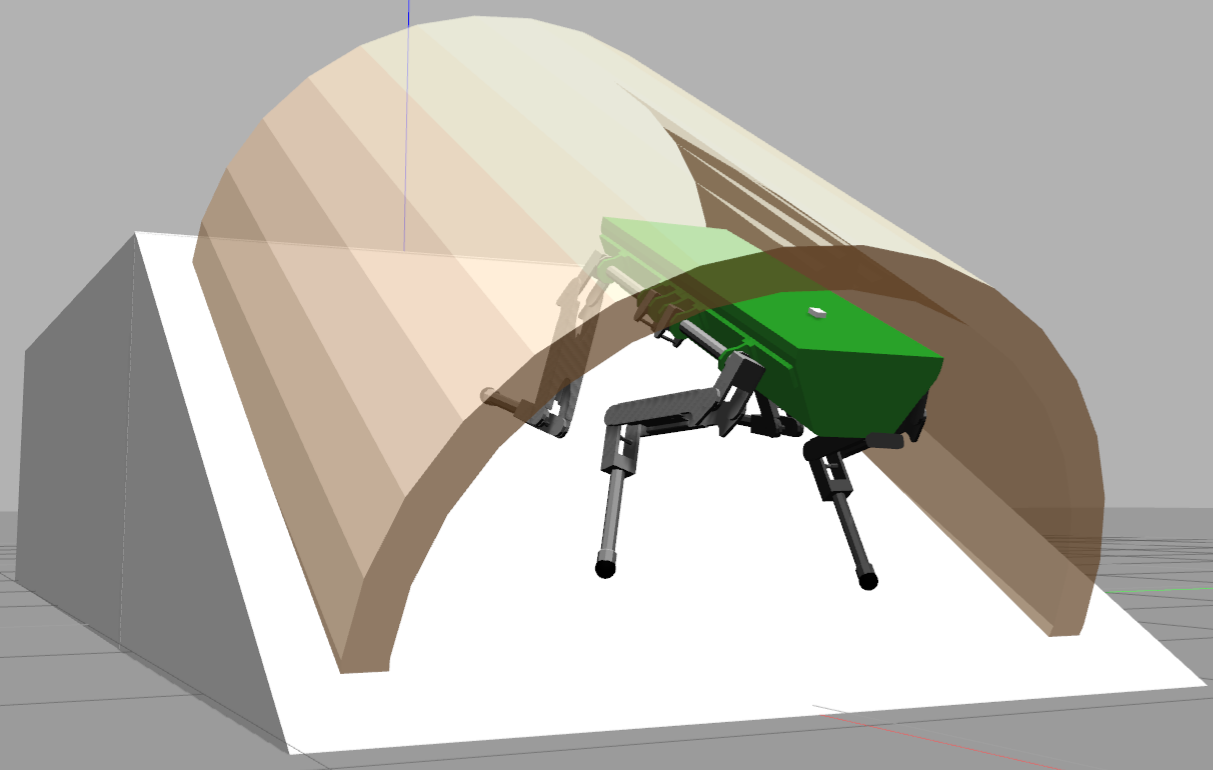}
\caption{\small Simulation of \gls{hyq} descending a challenging 30$\degree$ ramp with a 50 $cm$ tunnel (template 1). 
	The height of the HyQ robot is decreased from 53 $cm$ to 40 $cm$ in order to fit inside the tunnel. 
	A force-controllable rope (not shown in the figure) is attached to the back of the robot's trunk  to  compensate for gravity.}
\label{fig:gaz_bottom}
\end{figure}

\begin{figure}[]
\centering
\begin{subfigure}{7cm}
	\includegraphics[width=1\textwidth]{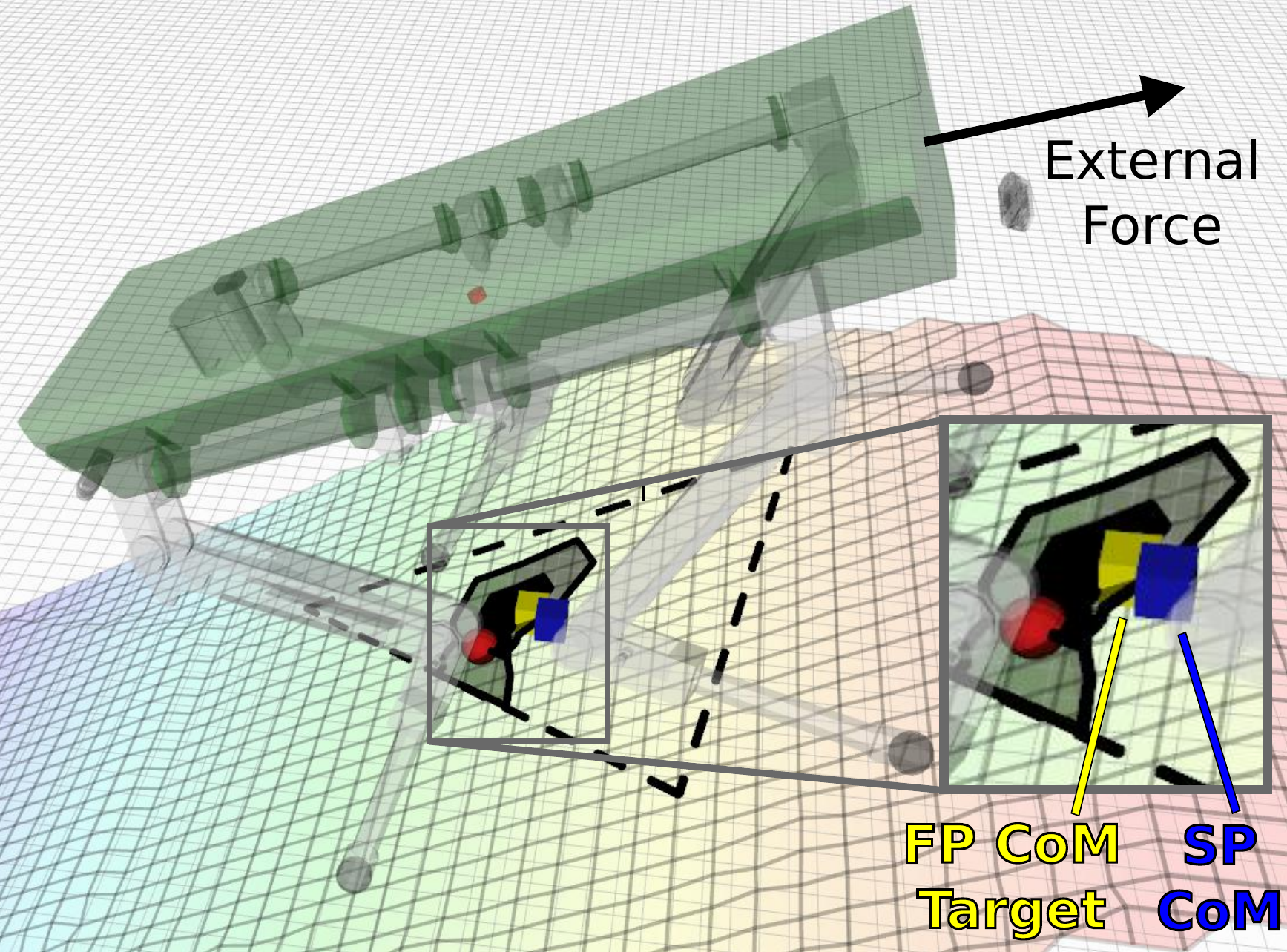}
	\caption{Middle of the \revisedtext{slope}}
\end{subfigure}\\
\begin{subfigure}{7cm}
	\includegraphics[width=1\textwidth]{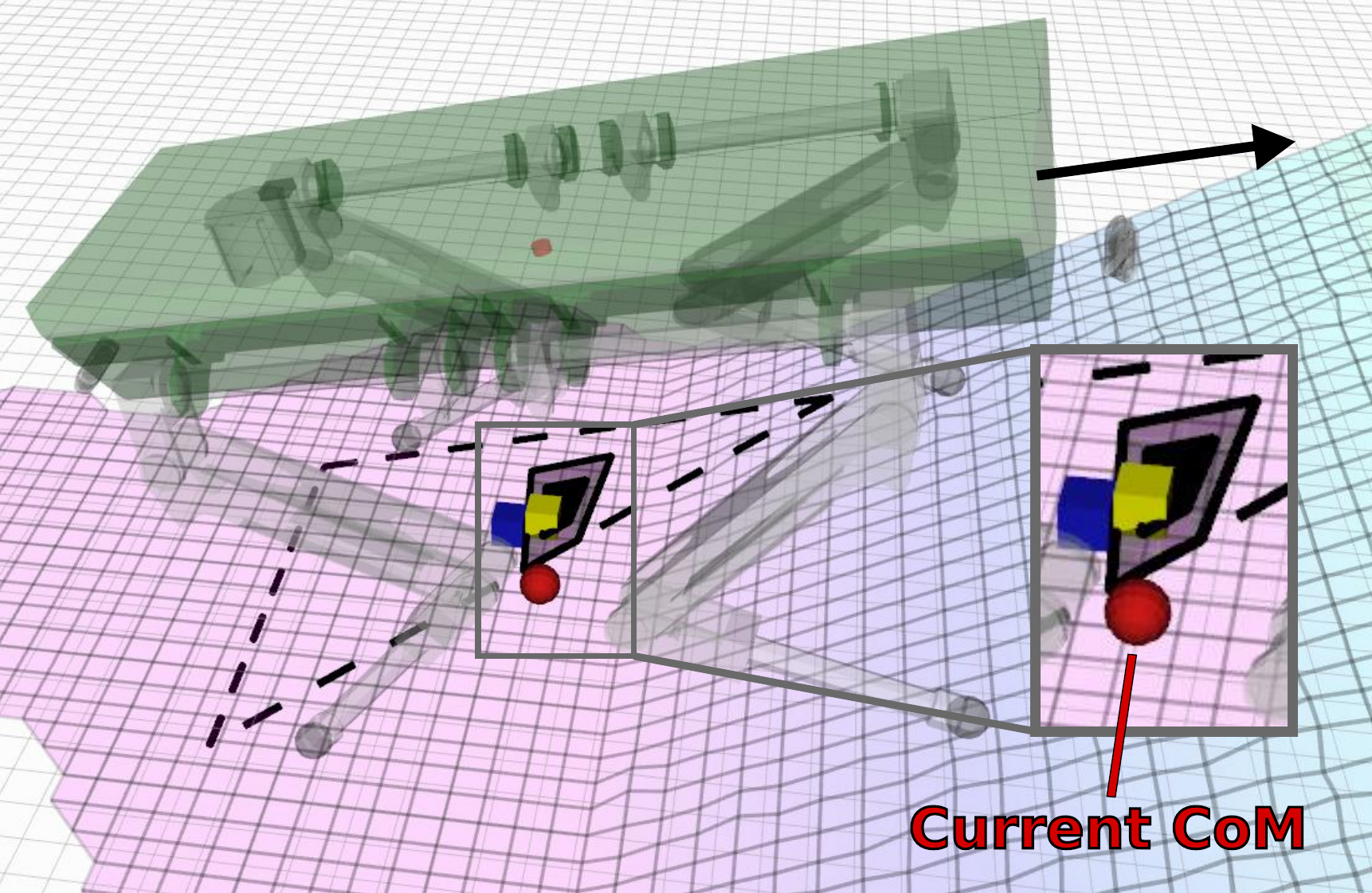}
	\caption{Bottom of the \revisedtext{slope}}
\end{subfigure}
\caption{\small Improved feasible regions and \gls{com} planning for two instances while descending the challenging tunnel in simulation (tunnel not shown in this figure).
	HyQ is heading to the left (downwards) while the external force due to the rope (black arrow) is applied  in a direction opposite to the motion.
	The regions shown above are for the future regions upon lift-off of the swing leg ($LF$ in the upper plot and $LH$ in the lower one):
	support regions (dashed),  improved feasible regions (grey), and the scaled  feasible regions (black).	
	Cubes represent the projection of the \gls{com} target based on \revisedtextthird{FP strategy} (yellow) and the \revisedtextthird{SP strategy} (blue), on the projection plane.
	Red sphere represents the projection of the current \gls{com}.
	This is out of the region because the robot  is still  moving toward the target, in the move-body phase (4 legs in stance). 
}
\label{fig:sim_duct}
\end{figure}

In this simulation, we assess the influence of 
an \textit{external wrench} acting on the robot, combined with a 
\textit{reduced} robot height necessary to walk 
in confined places. 
This challenging task consists of the \gls{hyq} robot descending a $30 \degree$ ramp while being  attached to a rope, to explore a low tunnel. 
This can be a typical scenario that a robot needs to face 
in oil rigs inspection assignments (see Fig. \ref{fig:gaz_bottom}).
A rope (not shown in the simulation software) connects the  back of the  robot to an anchor.
\revisedtext{The aid of the rope} results in regulated locomotion down the steep slope (\eg the same way a climber is \textit{rappelling} down a wall)\footnote{Experimentally, it is possible to attach the robot to an anchor where a torque-controlled electrically-driven hoist releases the rope while maintaining the required pulling force
(\ie the component of gravity force parallel to the sagittal axis of the trunk).}.
The role of the rope is to allow the contact forces to better satisfy friction constraints (\ie be more in the middle of the friction cones)
when walking on highly inclined terrains \cite{Bando2018}.
Indeed, in a slope with high inclination, the robot eventually creates a tangential force on the terrain that surpasses the friction force that is needed to prevent slippage.
An additional advantage of using a rope is that the robot can keep a more natural configuration, without the need to lean back or forth
to keep stability, thus keeping the joints in a more kinematically  advantageous configuration. 
As an additional difficulty, the restricted height of the tunnel places a risk of collision with the trunk of  \gls{hyq}. 
The robot is therefore forced to \textit{crouch walk} down the tunnel.
For this reason, we reduce the robot height from the default value of 53$cm$ to 40$cm$.\footnote{The robot height is defined as the distance between the \gls{com}  and the terrain plane along its  normal $\mathbf{n}$.}
This places the robot joints considerably close to their kinematic limits and in turn results in a restricted feasible region throughout the motion.
In addition, the feasible region will be shifted due to the influence of the external force (equivalent to 440 $N$ applied to the back of the robot) coming from the rope. %unextensible

The above-mentioned effects on the friction region  and on the feasible region  
can be seen in Fig. \ref{fig:sim_duct} for two instances in the simulation.
The regions are computed on the plane fitted through the stance legs \cite{Focchi2020}. 
This is parallel to the plane expressed by the robot trunk orientation where the \gls{com} planning is done.
In both situations, a shift in the friction and feasible regions, 
opposite to the external wrench on the robot, could be observed.
Furthermore, the low height imposed on the robot results in a big shrinkage of the feasible region.
Under these conditions, the  \gls{com} target (blue) planned with \revisedtextthird{the SP strategy} is located outside the region.
Conversely, the \gls{com} planner based on the improved feasible region \revisedtextthird{(FP)}, computes 
a feasible target (yellow) that is on the boundary of 
the \textit{scaled} feasible region and closest to the \revisedtextthird{SP} target.
It is interesting to remark that even though the friction region is shifted, 
thus giving the robot more freedom to lean forward if desired, 
the improved feasible region is inhibiting such courageous motions due to joint-torque restrictions and to the limited reachable region.
\subsection{Optimization of the Trunk Orientation on rough terrain}\label{sec:rough_terrain_exp}
\begin{figure*}[tb]
\centering
\begin{subfigure}{0.45\linewidth}
	\centering
	\includegraphics[width=0.75\linewidth]{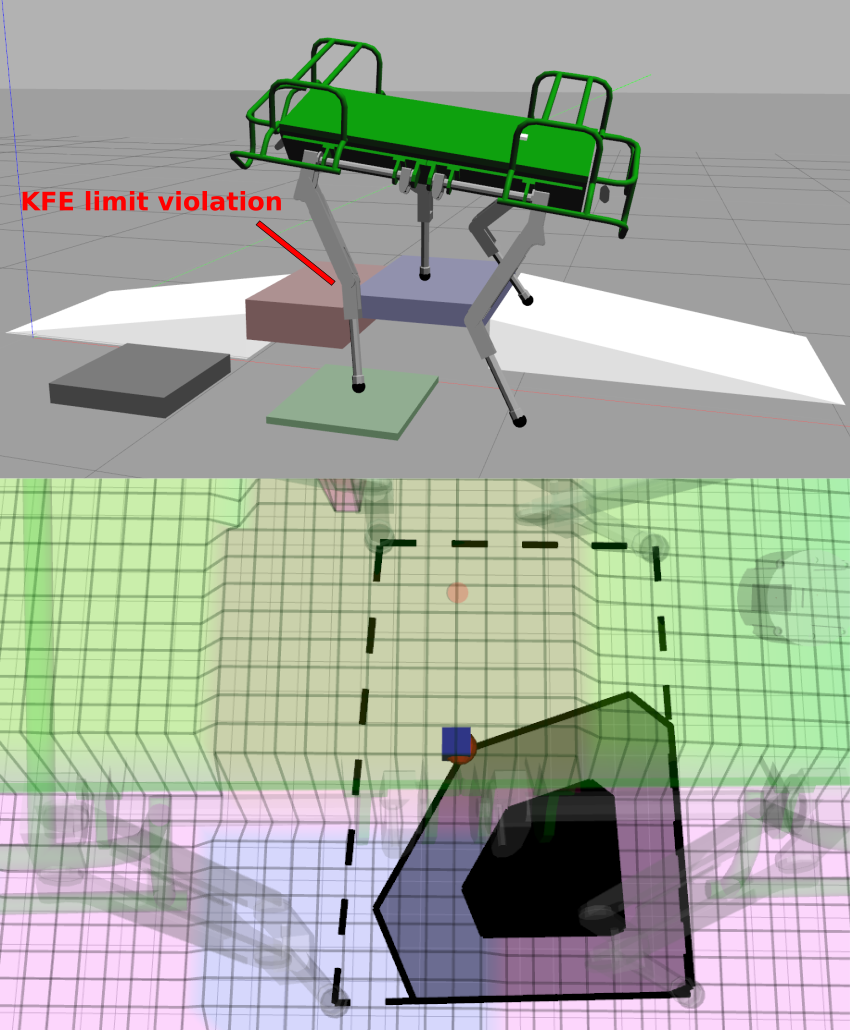}
	\caption{Using \revisedtextthird{Terrain-Based Planning}}
	\label{orient_heuristic}
\end{subfigure}
\begin{subfigure}{0.45\linewidth}
	\centering
	\includegraphics[width=0.75\linewidth]{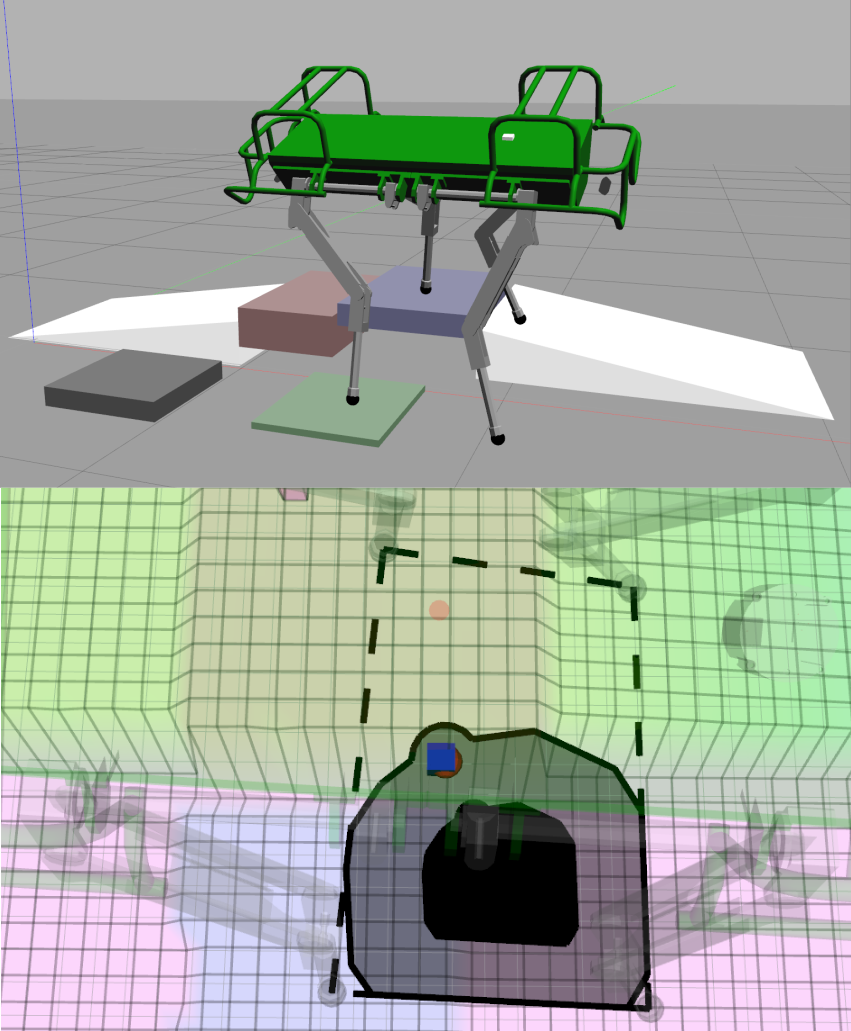}
	\caption{Using \revisedtextthird{Feasible Planning}}
	\label{orient_feasible}
\end{subfigure}
\caption{\small Simulation of HyQ forced near its kinematic limits while traversing a difficult non-coplanar terrain (Template 2). The robot configurations shown are at the end of a move-body phase.
	Realizing orientations based on  (a) the \revisedtextthird{TP strategy} and (b) based on the \revisedtextthird{FP strategy} results in different leg configurations (top). The resulting regions shown in the bottom plots are: friction regions (dashed),  feasible regions (grey), and the scaled  feasible regions (black). Large difference in the resulting  feasible regions can be seen, in turn affecting the feasibility of the \gls{com} trajectory (blue cube and red ball represent the projections of the \gls{com} target and the actual \gls{com}, respectively).
}
\label{fig:non_coplanar_terrain}
\end{figure*}

To illustrate the effectiveness of the orientation optimization strategy proposed 
in Section \ref{sec:orient_planning}, we test it separately from the 
\gls{com} planning strategy developed in Section \ref{sec:com_planning}. 
For this reason, the optimization of the orientation will be based on 
the \gls{com} target computed by the \revisedtextthird{SP} approach. 
As mentioned before, even if this does not  necessarily guarantee feasibility,
it allows us to compare clearly the improvements of the \revisedtextthird{FP strategy} over \revisedtextthird{the TP strategy}.
While climbing up a ramp, it is typical  to move 
the torso forward \cite{gehring2015,Focchi2020} in order to have the \gls{com} 
projection position closer to the middle of the support polygon, 
thus increasing the stability margin.
Therefore, aligning the trunk with the terrain inclination 
has the advantage\revisedtext{\sout{s}} of a superior  feasible region 
and consequently an ability to achieve a higher stability margin.
The case of the rough terrain shown in Fig. \ref{fig:non_coplanar_terrain}, 
is particularly challenging in terms of  kinematic limits. 
One of the legs can be forced to overly extend/retract during the 
move-body phase even though the other legs are possibly far from their limits.
In fact, adopting an orientation based solely on the \revisedtextthird{TP strategy} results 
in  infeasible trajectories in multiple 
locations of the terrain (Fig. \ref{fig:non_coplanar_terrain}(a) bottom).
The \revisedtextthird{TP} approach would not capture the difficulty given by the "lateral asymmetry" of this scenario. 
Indeed, it would result in a trunk with the hips being equally distant from the left and the right feet.
In the example shown, a pitch angle of $9.7\degree$ estimated averaging terrain plane), is selected by the \revisedtextthird{TP} approach. 
This results in a hyper-extension of the \gls{rh} leg and a 
kinematic violation at the \gls{kfe} joint (Fig. \ref{fig:non_coplanar_terrain}(a) top).
Note that since we model the kinematic limits 
in our simulator, the \gls{com} will not be allowed to go out of the boundary of the region. 
The same \gls{com} trajectories could instead be feasible if the orientation is planned based on the \revisedtextthird{FP strategy}, with an optimized pitch angle of $-0.3\degree$  (see Fig. \ref{fig:non_coplanar_terrain}(b)).
The optimized pitch angle maximizes the distance of the trajectory, from the boundary of the region (\ie the margin), as well as the area of the region, thus resulting in a safer joint configuration.

\section{Experimental Results} 
\label{sec:experiments}

\subsection{Walk in cluttered environment}
\begin{figure}[tb]
\centering
\includegraphics[width=0.5\textwidth]{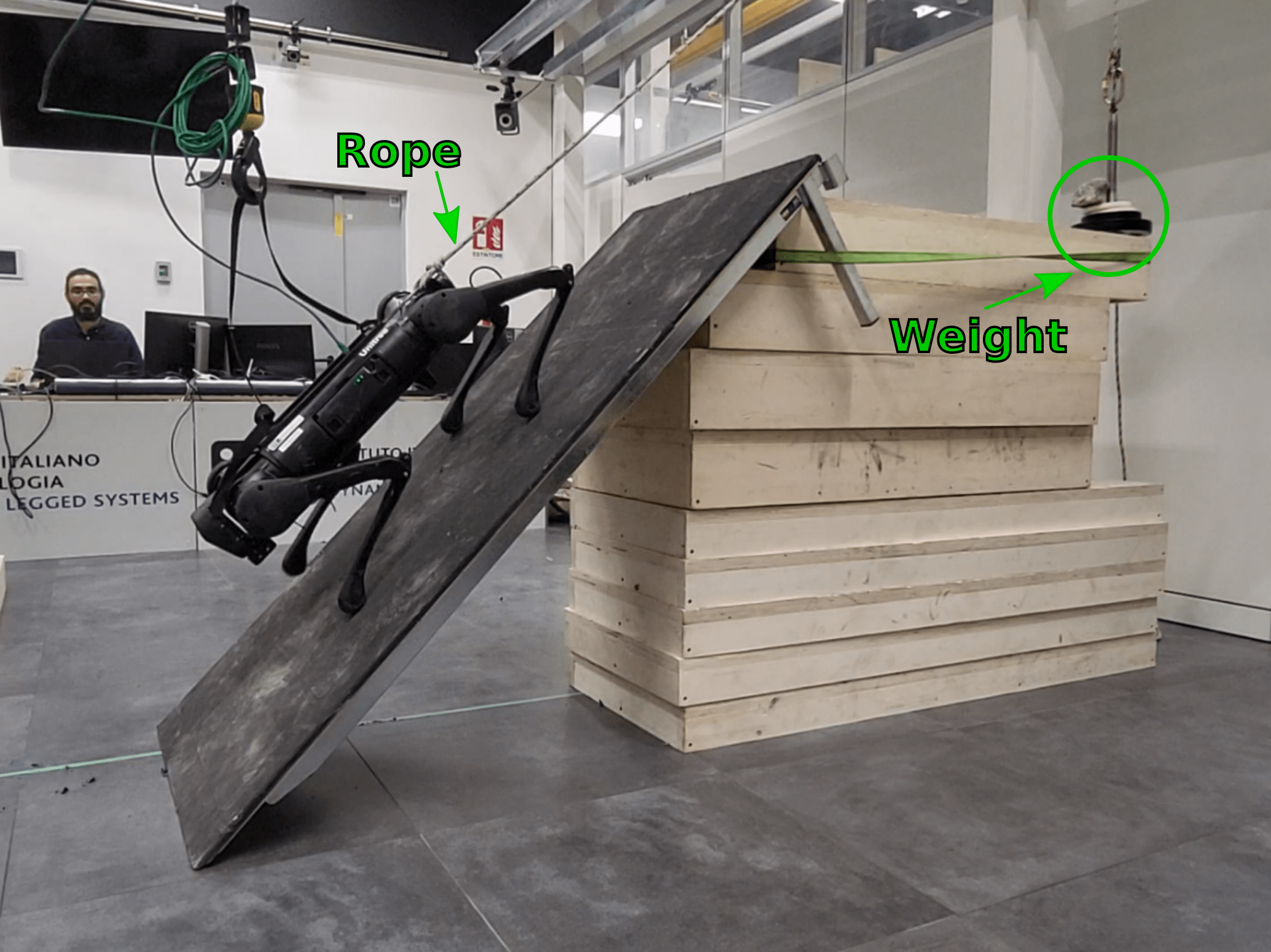}
\caption{\small Aliengo descending a 45$\degree$ ramp with a reduced robot height of 20 $cm$ (similar to the scenario shown in simulation). 
	A rope is attached to the back of the robot's trunk  to  compensate for gravity through a counterweight.}
\label{fig:ramp_experiment}
\end{figure}
We implement the simulation example shown in Section \ref{sec:simulations}.A on real hardware using the robot Aliengo (for safety reasons) as shown in Fig. \ref{fig:ramp_experiment}. The robot is commanded to walk down a steeper slope of 45$\degree$ with a robot height of 20 $cm$. A pulley, rope and counterweight (a mass of about 10 kg was used) are used to introduce the external force needed to pull the robot backwards. The kinematic limits of Aliengo are virtually lowered to simulate that of hydraulic robots like HyQ at low robot heights.

\begin{figure}[h!]
\centering
\includegraphics[width=0.485\textwidth]{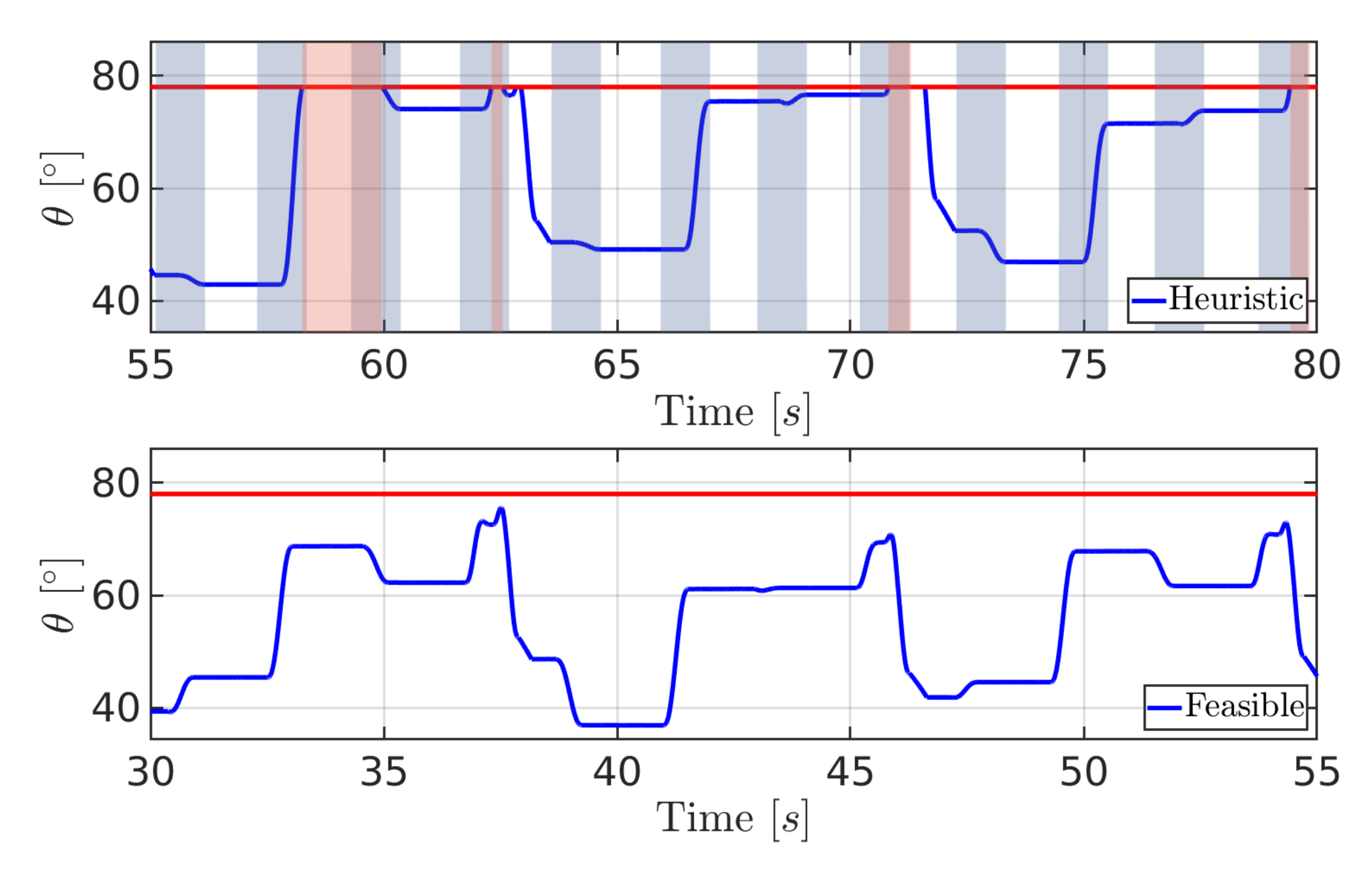}
\caption{\small Experimental results showing the Right-Hind \gls{hfe} joint trajectory of Aliengo during the ramp descent. 
	\revisedtextthird{SP strategy} (above): the knee starts to hit the virtual kinematic 
	limit (red line) during the \textit{move body} phases (shaded blue). The violations are in shaded red. \revisedtextthird{FP strategy} (bottom): no kinematic limit violations are observed. }
\label{fig:exp_low_height_limits}
\end{figure}

The plots of the \gls{kfe} joint trajectory during the 
experiments are reported in Fig. \ref{fig:exp_low_height_limits}. A \gls{com} \revisedtextthird{target based on the SP strategy} would result in multiple violations of the kinematic limits
(upper plot) while the one based on the \revisedtextthird{FP strategy} has no violations (lower plot).

\begin{figure}[tb]
\centering
\includegraphics[width=0.485\textwidth]{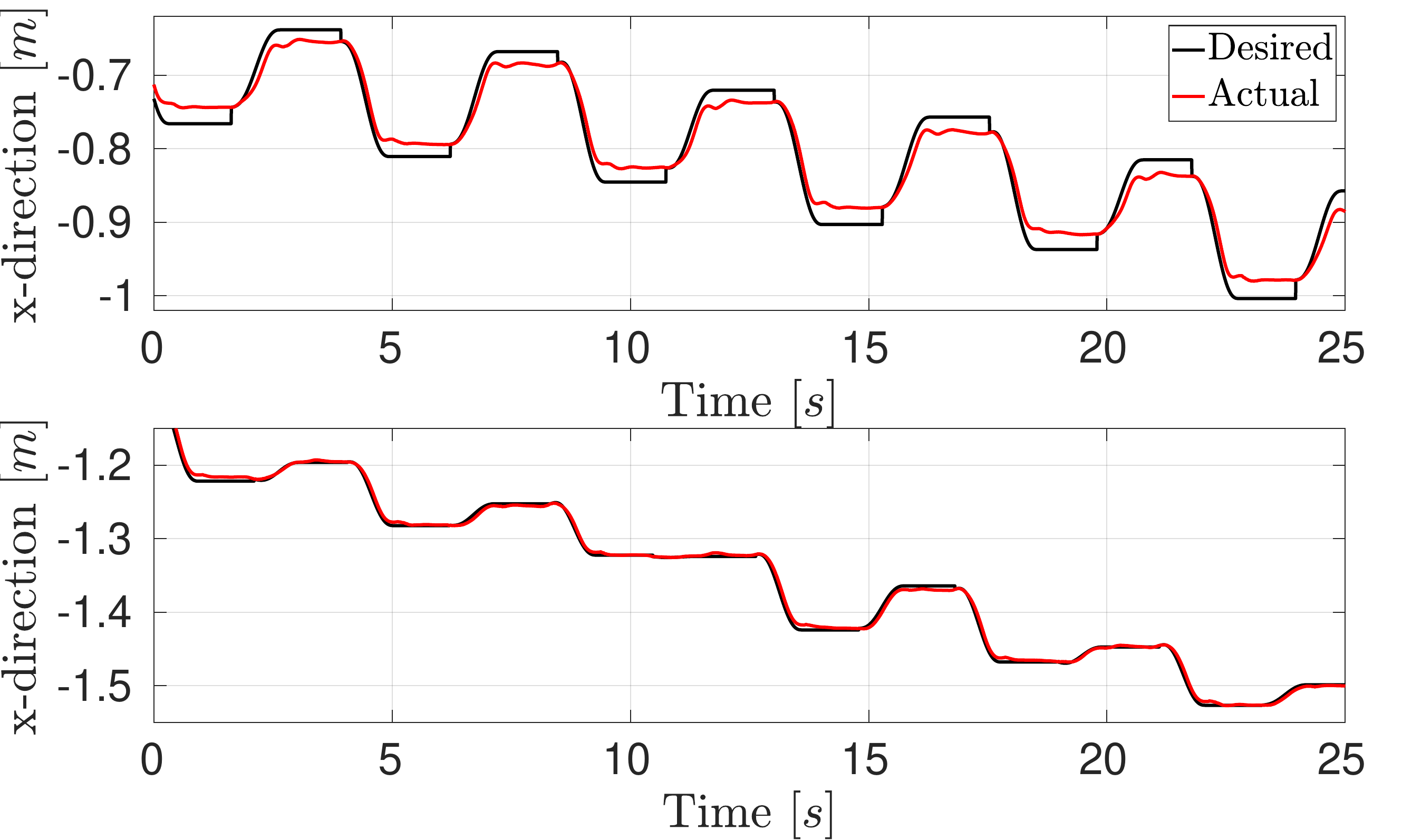}
\caption{Experimental results showing the \gls{com} position tracking of \gls{hyq} in the $x$ direction. 
	A deterioration can be seen with the \revisedtextthird{SP strategy} (upper plot) 
	due to the joint kinematic limit violations while good tracking is observed \revisedtextthird{with the FP strategy} (bottom plot).}
\label{fig:exp_low_height_tracking}
\end{figure}

Additionally, to show the effect kinematic violations can have on the performance of the robot, we perform experiments with the 90 kg \gls{hyq} robot platform walking on flat ground with a reduced height of 43 $cm$. Fig. \ref{fig:exp_low_height_tracking} shows that such  kinematic violations result in a deterioration of the tracking of the \gls{com} trajectory \revisedtextthird{computed using the SP strategy} as opposed to the \revisedtextthird{FP strategy}.


\revisedtext{\subsection{Dynamic Motions}
In this section we show an experiment with the 21 kg robot Aliengo performing a dynamic trotting gait developed in an earlier work \cite{barasuol13icra}.
The improved feasible region for the motion is computed, encompassing the dynamic effects as explained in Section \ref{sec:dynamics}.
The purpose of this is two-fold: 1) demonstrate the inertial effects that dynamic motions have on the improved feasible region, and 2) to show how the solution we proposed in Section \ref{sec:degenerate}  
to deal with  "degenerate" regions works effectively. We report the results in the accompanying video showing that the \gls{com} projection remains
inside the feasible region. }

\section{Conclusion}
\label{sec:conclusions}
In this paper we introduced an improved version of the feasible region 
presented in our previous work \cite{orsolino19tro}.
The feasible regions are intuitive yet powerful and computational efficient tools to 
plan feasible trajectories for a reference point of the robot (\eg the \gls{com}).
The original feasible region, that was defined as the set of 
\gls{com} positions where a robot is able to maintain static equilibrium
without violating friction and actuation limits, 
was extended to take into account also 
kinematic limits (through the newly defined \textit{reachable region}) and the presence of external wrenches acting 
on arbitrary points of the robot. This, along with projecting the region on arbitrary plane inclinations that are consistent 
with the planning intention of the user, offers the opportunity to employ the 
proposed motion planning strategy to new possible applications. One such application has been demonstrated in this work, i.e. rope-aided locomotion, while the same approach can be applied to load-pulling/pushing applications.
To include the dynamic effects of motion we relaxed the quasi-static assumption in the iterative projection algorithm.

Furthermore, we proposed a planning  strategy
that utilizes the improved feasible region to design feasible \gls{com} and trunk
orientation trajectories. 
We validated the capabilities of the improved
feasible region and the effectiveness of the proposed planning
strategy on challenging simulations and experiments with the \gls{hyq} and Aliengo
robots and we compared our results to a
previously developed approach \cite{Focchi2020} that 
is not able to formally guarantee the kinematic feasibility of its trajectories. \revisedtext{We validated the extensions of the feasible region to be compatible with dynamic motions using experiments with the Aliengo robot and demonstrated the effect that the height and orientation of the robot have on the reachable region.}

%future works
\revisedtext{As future works, we intend to incorporate the feasible region in an centroidal
momentum MPC controller. This will result in a CoM constraint that is suitable for non coplanar contacts and would add descriptiveness to the MPC formulation (employing a reduced model) with joint-based constraints (e.g. torque and/or kinematic constraints that are embedded in the region).}
Other ongoing works are focused on speeding up the computation of the 
region increasing its accuracy in the vicinity of the  direction of motion. 
This would allow us to only refine (or even only compute) the parts of the  
feasible region that are relevant to the locomotion direction.
%idea of cost
\appendix
\label{sec:appendix}
This appendix provides more theoretical analysis on some aspects of the reachable and improved feasible region, as well as provide additional implementation details to the interested readers.

\subsection{Reachable region and CoM workspace}
\begin{figure}[tb]
\centering
\includegraphics[width=0.4\textwidth]{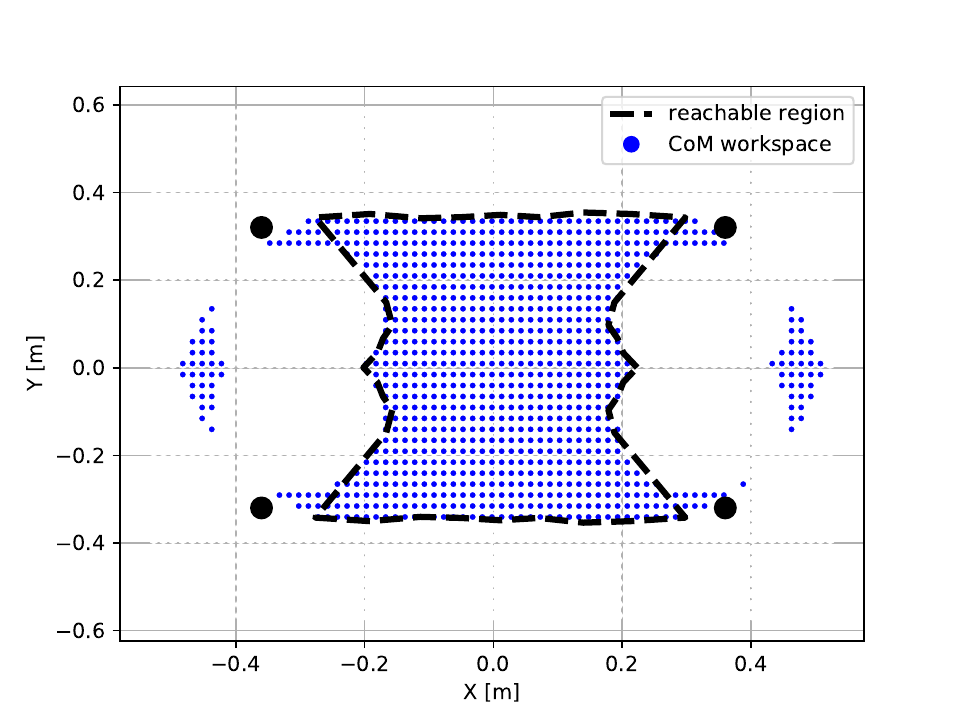}
\caption{\small Comparison between the CoM workspace (ground-truth \revisedtextthird{blue}) and \textit{reachable region} (black dashed) for \gls{hyq} at 0.49 $m$ CoM height.}
\label{fig:full_workspace}
\end{figure}

It is useful to further illustrate the effect of the ray casting algorithm on the produced reachable region as compared to the full workspace of the CoM. \revisedtext{The full workspace can be comprised of disjoint sets (\eg \cite{stewart_numerical}) 
which would not be captured by the algorithm. Figure \ref{fig:full_workspace} shows the \gls{com} workspace for a height of 0.49 $m$ for the HyQ robot, where the green and red points show the kinematically feasible and infeasible locations, respectively, obtained using a brute force approach where each point in a two-dimensional grid (grid point distance of 2.5 cm)  was tested and  marked accordingly.}

\revisedtext{As evident in Fig. \ref{fig:full_workspace}, a special case can arise at some configurations where the workspace is non-convex and some disjoint sets appear. Given the nature of ray casting algorithms, in such cases the method only determines the reachable region in the range of rays casted from an initial point. The black dashed boundary in the figure shows the output of the algorithm when  started from the center of the workspace. In fact, the disjoint regions on the right and left sides are undetected by the algorithm.
A solution for this can be to start the algorithm from different parts of the workspace and attempt to rebuild the workspace. However this introduces needless complexity because disjoint regions are nevertheless infeasible for planning, given that no continuous trajectory can be constructed.
}

\subsection{Regions intersection}
The \textit{improved feasible region} can simply be obtained by intersecting the \textit{feasible region} with the \textit{reachable region}. This is in contrast with the case of attempting to obtain the \textit{feasible region} $\mathcal{Y}_{fa}$ by the simple intersection of the friction region $\mathcal{Y}_{f}$ and the actuation region (centroidal mapping of joint-torques) $\mathcal{Y}_{a}$ as explained in \cite{orsolino19tro}.
In general, since $\mathcal{C}$ and $\mathcal{A}$ are defined on the same space, the intersection of the two sets (\eg stacking both friction and joint-torque constraints) must be carried out first before projecting the resulting set.
The converse is inaccurate since the intersection and projection operators do not commute.
In the case of the reachable region the constraints are defined not on contact 
forces but on joints angular positions, so this issue does not exist.

\subsection{Non-convex scaling}
\revisedtext{Scaling a non-convex polygon through an \textit{affine} transformation with respect 
to a reference point (e.g., the Chebyshev center or the centroid) could result in a scaled region with parts outside the original one.
On the other hand, inward \textit{polygon offsetting} algorithms guarantee that the scaled polygon always remains inside the original one.
One downside of this algorithm is that the scaled polygon can suffer from topological changes (e.g., some edges might contract until they vanish \cite{offset}).
Furthermore, the scaling during the offsetting procedure is defined by a distance. The centroid based scaling, on the other hand, characterizes the scaling in terms of a percentage. This allows the algorithm to be directly scalable and consistent with robots of different dimensions.}
Although offsetting non-convex polygons is still a hard problem in itself,
\cite{shrinking} proposed a solution for non-convex polygons.

\begin{figure}[tb]
\centering
\includegraphics[width=0.4\textwidth]{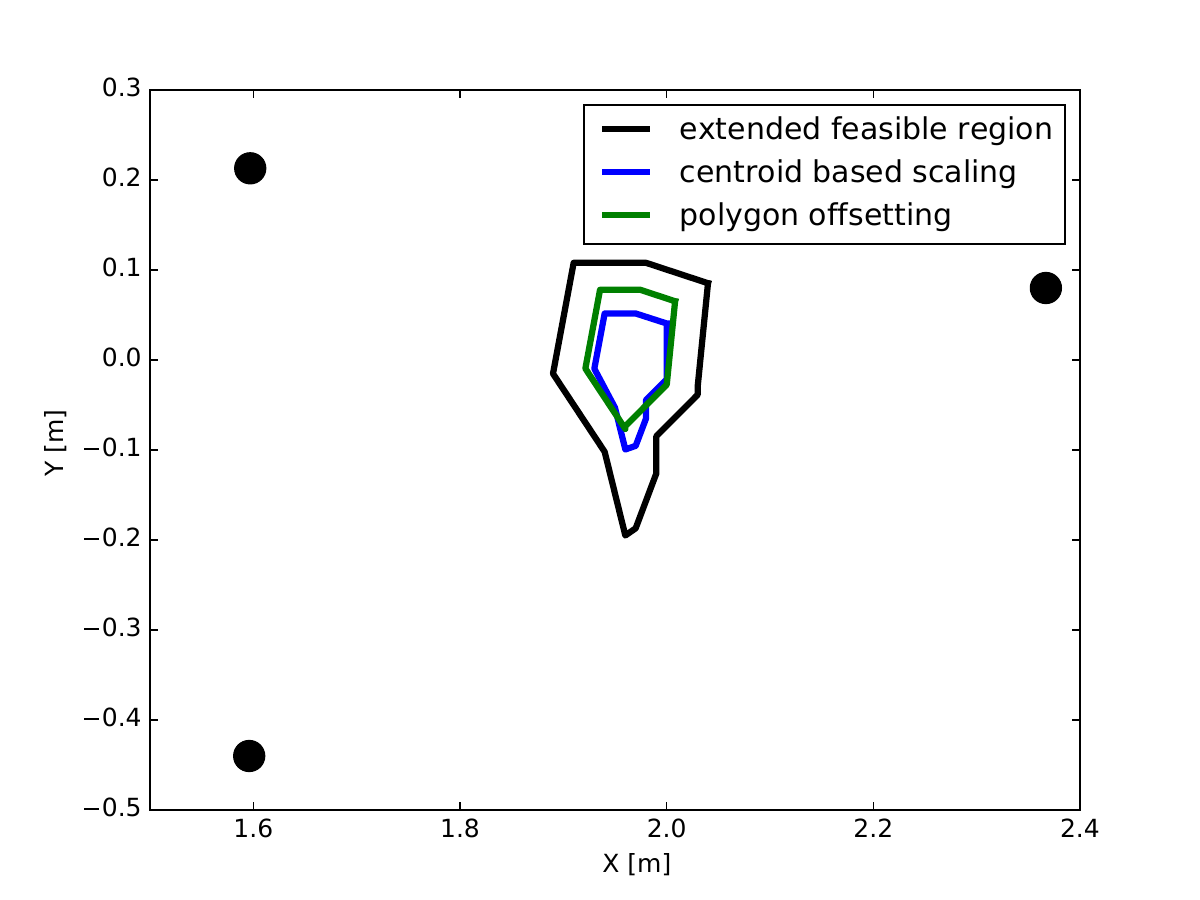}
\caption{\small Comparison between different methods of scaling the (non-convex) improved feasible region: scaling based on the centroid and scaling using polygon offsetting.}
\label{fig:polygon_scaling}
\end{figure}

\revisedtext{In Fig. \ref{fig:polygon_scaling} we compare the result achieved using the polygon offsetting algorithm to the output of the affine scaling with respect to the centroid. An improved feasible region (shown in black) is computed for a three contacts phase with a height of 0.37 $m$ for the HyQ robot.
We set a scaling factor $s = 0.5$  for the affine scaling (black) and an offset of $r = 0.03$ $m$ for the polygon offsetting (green).
A small difference in the area of the scaled polygons and a slight shift can be observed, 
with the offsetted polygon changing its topology due to the  already small size of the original polygon.
We chose to utilize the affine scaling for the direct scalability and its efficiency (an average computation time of 0.005 $ms$ as compared to 5 $ms$ for the polygon offsetting). To guarantee that the scaled region would strictly be a member of the original one,
an additional efficient step of using the intersection of the original and the scaled region can be performed. }

\small
\section*{Acknowledgements}	
We would like to thank Victor Barasuol for the valuable advice during the experiments and review process of this paper. We would also like to thank Shamel Fahmi, Chundri Boelens, and all the members of the DLS lab for the help 
provided in the development of this work.

\bibliographystyle{IEEEtran}
\bibliography{references/references.bib}

\begin{minipage}[t]{0.45\textwidth}
\begin{IEEEbiography}[{\includegraphics[width=1in,height=1.25in,clip,keepaspectratio]{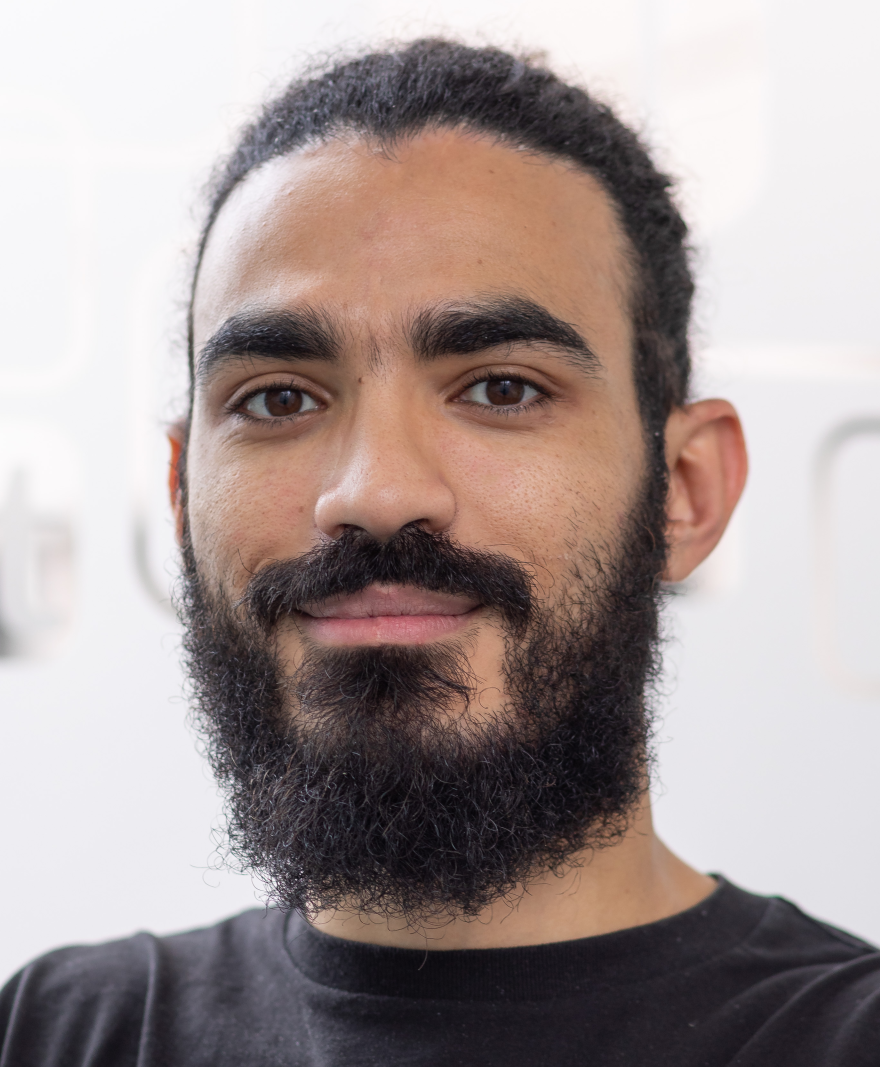}}]{Abdelrahman Abdalla}
	was born in Cairo, Egypt. He received the B.Sc. degree in
	mechatronics from the German University in Cairo,
	Cairo, Egypt, in 2017, the M.Sc. degree in control engineering from the Sapienza Università, Rome,
	Italy, in 2020, and the Ph.D. degree in
	advanced and humanoid robotics from the Italian
	Institute of Technology, Genoa, Italy, in 2021.
	In December 2021, he joined the Dynamic Legged
	Systems Lab, Istituto Italiano di Tecnologia for the
	Ph.D. degree.
	His Ph.D. focuses on wrench-based locomotion planning and the development of efficient locomotion feasibility criteria.
\end{IEEEbiography}

\end{minipage}
\begin{minipage}[t]{0.45\textwidth}
	\begin{IEEEbiography}[{\includegraphics[width=1in,height=1.25in,clip,keepaspectratio]{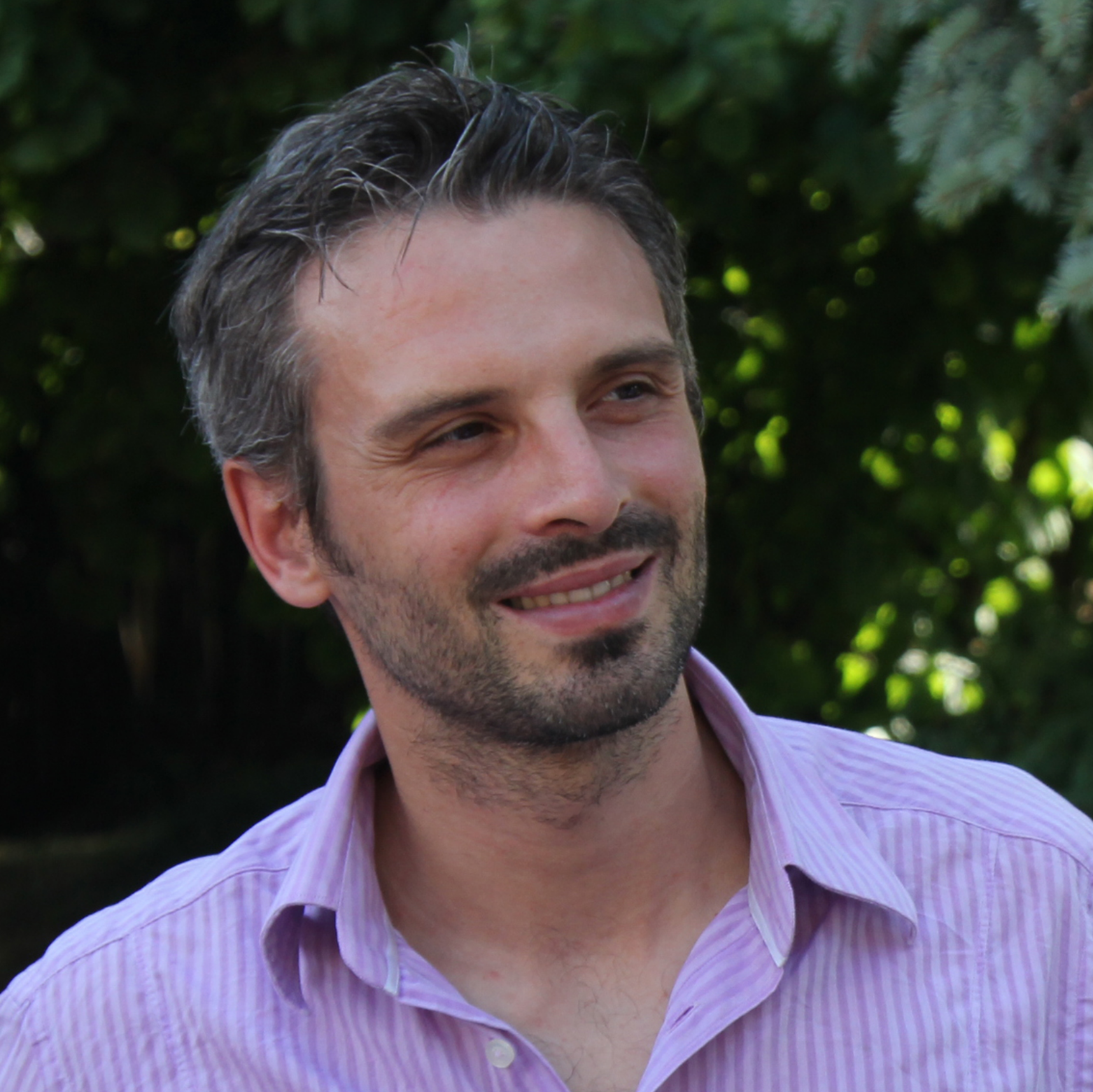}}]{Michele Focchi}
		was born in Rimini, Italy, in 1980. He received both the B.Sc. and the M.Sc. degree in control systems engineering from Politecnico di Milano, Italy in 2007.
		He received the Ph.D. degree in robotics from Istituto Italiano di Tecnologia (IIT), Genova, Italy, in 2013. 
		
		From 2014 to 2021 he worked as a researcher at IIT. His research has been concerned with  the software development of planning and control strategies for quadruped robots. On 2022 he joined University of Trento, Italy where he works as a lecturer. Currently his research interests are focused on pushing the performances of quadrupeds in traversing unstructured environments, by using optimization-based techniques and devising novel robotic platforms.
	\end{IEEEbiography}
	\begin{IEEEbiography}[{\includegraphics[width=1in,height=1.25in,clip,keepaspectratio]{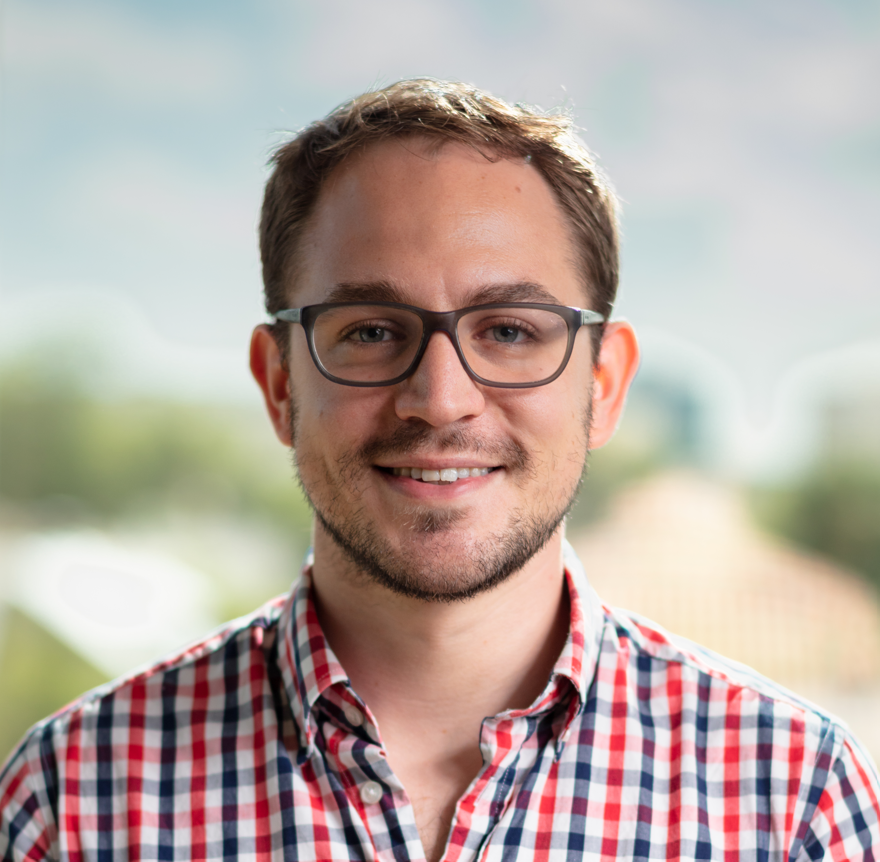}}]{Romeo Orsolino}
		completed his Ph.D. in Bioengineering and Robotics from Istituto Italiano di Tecnologia (IIT) in 2019, where he worked at the DLS lab focusing on motion planning for legged locomotion in rough terrains. He then joined the DRS group at the Oxford Robotics Institute, University of Oxford, as a postdoctoral researcher where he pursued his research on multi-contact motion planning, optimal control, dynamics and perception.
		
		Since December 2020, Romeo has moved to industry where he works as a robotics research engineer on R\&D projects to increase the use of advanced robotics in the automotive and energy sectors. 
	\end{IEEEbiography}
\end{minipage}
\hfill
\vspace{9.0cm}
\begin{minipage}[t]{0.45\textwidth}
	\begin{IEEEbiography}[{\includegraphics[width=1in,height=1.25in,clip,keepaspectratio]{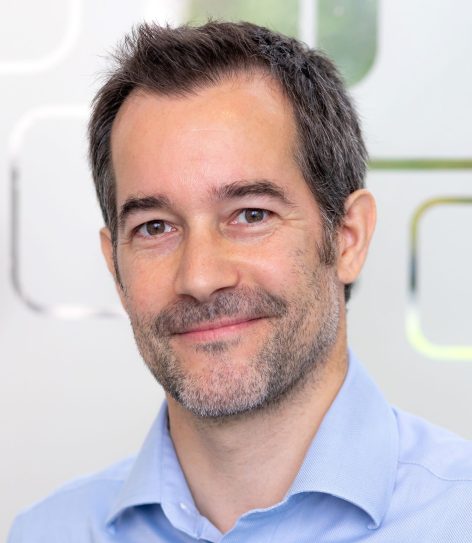}}]{Claudio Semini}
		(Member, IEEE) received the M.Sc. degree in electrical engineering and information technology from ETH, Zurich, Switzerland, in 2005, and the Ph.D. degree in humanoid technologies from the Istituto Italiano di Tecnologia (IIT), Genoa, Italy, in 2010.
		During his doctorate, he developed the hydraulic quadruped robot HyQ and worked on its control.
		
		Since 2012, he has been leading the Dynamic Legged Systems Lab, IIT. He is/was the coordinator/partner of several European Union, National, and Industrial projects (including HyQ-REAL, INAIL Teleop, Moog@IIT joint lab, ESA-ANT, RAISE, FAIR, etc.). He is the author and co-author of over 100 journal and conference publications. His research interests include the design and control of legged robots for real-world operations, locomotion, and hydraulics.
		
		Dr. Semini is a Co-Founder of the Technical Committee on Mechanisms and Design of the Robotics and Automation Society.
	\end{IEEEbiography}
\end{minipage}
\end{document}